\documentclass{article}
\usepackage{iclr2027_conference,times}
\usepackage{amsmath,amsfonts,bm}

\def\eqref#1{equation~\ref{#1}}

\def\1{\bm{1}}

\DeclareMathAlphabet{\mathsfit}{\encodingdefault}{\sfdefault}{m}{sl}
\SetMathAlphabet{\mathsfit}{bold}{\encodingdefault}{\sfdefault}{bx}{n}

\usepackage{hyperref}
\usepackage{url}
\usepackage{float}
\usepackage{graphicx}
\usepackage{array}
\usepackage{tabularx}
\usepackage{booktabs}
\usepackage{subcaption}
\usepackage{multirow}
\usepackage{xcolor}
\usepackage[most]{tcolorbox}
\usepackage{enumitem}
\usepackage{wrapfig}
\title{Not All Pairs Align: Caption-Guided Audio-Visual Contrastive Learning}
\author{Omar Emara\textsuperscript{1}\quad Nan Lu\textsuperscript{1} \quad Dima Damen\textsuperscript{1} \quad Hazel Doughty\textsuperscript{2} \\ \textsuperscript{1}University of Bristol \quad \textsuperscript{2}Leiden University }
\date{}

\definecolor{deltagrey}{gray}{0.45}

\iclrfinalcopy

\begin{document}

\maketitle

\begin{abstract}

We address the problem of learning audio-visual representations from co-occurring modalities.
Multimodal contrastive learning typically assumes that paired observations provide shared views of the same underlying information. While this may be true for datasets collected in constrained environments, it does not always hold. For instance, co-occurring audio and visual signals can differ substantially in their content due to off-screen sounds. Nevertheless, standard contrastive objectives treat all paired samples as equally aligned. In this work, we propose \textbf{C}aption \textbf{G}uided \textbf{A}lignment \textbf{A}ware \textbf{L}oss (\textbf{CGAAL}), a model-agnostic objective that leverages modality-specific captions to estimate semantic agreement between audio and video. CGAAL uses these agreement scores to weight the contribution of each positive pair. A warm-up period and gradual thresholding strategy progressively introduce this alignment-aware training signal. We demonstrate the generality of CGAAL by applying it to three contrastive baselines, consistently improving their performance on audio-visual retrieval.

\end{abstract}

\vspace{-0.8em}
\section{Introduction}
\vspace{-0.6em}

Multimodal contrastive learning has emerged as an effective approach for learning transferable representations from naturally paired data, e.g., image--text and audio--video pairs \citep{CLIP, CLAP, alayrac2020self, chen2021multimodal}. By treating paired observations as positives and unpaired ones as negatives, these methods learn a shared embedding space without labels. Their success is often explained by the multiview redundancy assumption: each modality shares enough task-relevant information to learn useful representations \citep{multiview_red}.

\begin{figure}[b]
    \centering
    \includegraphics[width=\textwidth]{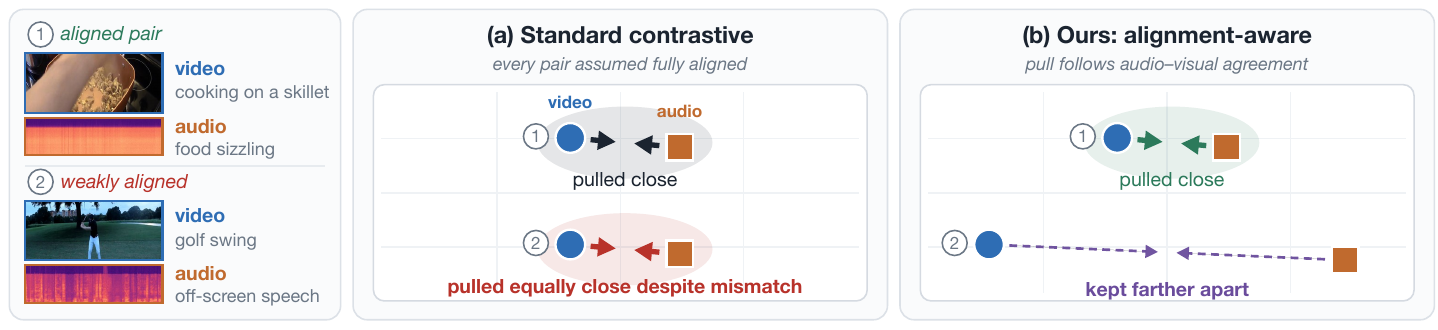}
    \vspace{-2em}
    \caption{\textbf{Alignment in audio-video contrastive learning}. Standard contrastive methods assume strong semantic correspondence between modalities of a given sample, encouraging alignment even when the modalities share little semantic content. This can encourage the model to associate audio and visual content that does not correspond. We thus propose CGAAL which determines how strongly to pull the modalities' embeddings of a sample according to a pair's semantic agreement.}
    \label{fig:teaser}
\end{figure}

This assumption is particularly problematic for audio and video. Although audio-visual signals may co-occur in the same video clip, they need not be semantically aligned.
For example, a video showing someone playing golf while its audio contains speech unrelated to the game. Under the multiview redundancy assumption, standard contrastive learning nevertheless encourages strong alignment of all pairs, introducing a potentially misleading training signal (see Figure~\ref{fig:teaser}).

Recent work challenges the assumption that multimodal observations carry only redundant information, distinguishing shared information  from information that is unique to one modality or that emerges only through their interaction \citep{CoMM, infMasking}. These methods address this by learning a single representation that captures shared, unique, and synergistic components, and have shown clear gains over purely redundancy-based objectives. We approach the same problem from a different angle. Rather than explicitly capturing unique and shared information, we act on the alignment objective itself addressing how strongly modalities in a pair should be aligned.

We introduce \textbf{CGAAL}\footnote{We pronounce `CGAAL' as `Seagaal' for simplicity.}, a \textbf{C}aption \textbf{G}uided \textbf{A}lignment \textbf{A}ware \textbf{L}oss, which incorporates an estimate of each pair’s semantic correspondence into contrastive training. From modality-specific captions describing the audio and visual content, we use an LLM to estimate how closely the two modalities semantically align. These scores weight the positive pairs during contrastive training, placing less emphasis on pairs that share little semantic content (see Figure \ref{fig:teaser}). We combine this weighting with a curriculum learning inspired approach to incorporate the alignment scores progressively. Importantly, CGAAL requires neither captions nor an LLM at inference time, and can be integrated into different contrastive learning frameworks. Our contributions are:

(1) We identify variable semantic correspondence as a fundamental limitation of audio-visual contrastive learning and show that methods designed to distinguish shared from unique, and synergistic information do not resolve this issue for cross-modal retrieval.
\newline
(2) We propose to use modality-specific captions as a guide for semantic alignment between modalities in each sample by incorporating an alignment score into the training loss via a weighting scheme, adjusting how strongly paired embeddings are pulled together.
\newline
(3) We demonstrate our method's ability to improve audio visual retrieval performance across a range of contrastive learning models and show zero-shot capabilities on three unseen datasets.

\vspace{-0.4em}
\section{Related Work}\label{sec:related_work}
\vspace{-0.6em}

\textbf{Multimodal contrastive learning} uses naturally paired data to align representations from different modalities, typically through an InfoNCE loss \citep{InfoNCE}. A prominent example is CLIP~\citep{CLIP}, which contrasts image-text pairs to learn a shared embedding space. Similar contrastive objectives have since been applied to other modality pairings, such as audio-text~\citep{CLAP}, audio-visual~\citep{Parida_2026_CVPR}, video-text~\citep{text_video} and point cloud-text~\citep{Paul_2026_CVPR}, as well as extended to jointly align three or more modalities~\citep{VAST, VALOR, everything_at_once, guzhov2022audioclip, InternVideo2, ImageBind}.  These approaches apply contrastive alignment to global representations of each input. Other work instead considers alignment at finer granularities. For instance, video-text retrieval approach X-CLIP ~\citep{x_clip} computes similarities between videos, frames, sentences and words. More recently, FDCA \citep{yue25finecvr} first separates the text at sentence and token level and then aligns them to the video. Audio-video retrieval approaches have also considered finer-grained alignment: CAV-MAE \citep{cav_mae} aligns visual frames with temporally localised audio segments, while \cite{Fan2026} improve spatial alignment through multiscale spatial attention that aligns audio with particular objects and events.

Despite differing in the number and granularity of the modalities they align, these approaches all treat naturally co-occurring modalities as positive pairs regardless of how much semantic content the modalities share. In contrast, we estimate a modality alignment score and use this to control how strongly each pair should be aligned.

\textbf{Learning with partial cross-modal correspondence.}
Several approaches have attempted to relax the assumption that paired samples are equally aligned.
Similar to our work, soft-weighted objectives have been used by~\citet{AVID, CWCL, CellCLIP, PhenoMolecular}.
However, to obtain the soft-weighting, these approaches first train a model under the multiview-redundancy assumption, assuming all pairs are strongly aligned. The similarities in this embedding space then act as the soft-weights to guide the partial cross-modal correspondence.
We argue that this approach is fundamentally problematic as the model trained with the multi-view redundancy assumption will have already lost much of the nuance in a pair's alignment.

Other related works assume partial alignment in image-text paired data where only a part of the image should be aligned with certain words in the caption~\citep{https://doi.org/10.48550/arxiv.2212.07699, 11095210,huang2021gloria}.
All these works assume that some alignment between part of the image and words in the caption exists.
This is not the case in audio-visual data where at times the audio and the vision do not align at all (random music or speech that is not relevant to the visuals).
Differently, we rely on text to better approximate the alignment of the vision and audio to weight the positive pairs rather than relying on similarities based on the modalities' embeddings.

Closer to our work is Sounding Actions~\citep{SoundingActions} which uses text to establish a consensus signal for audio-visual alignment.
Their focus is on non-audible actions, i.e. actions for which audio is silent.
They learn to align audio and vision when both correspond to the action described in the visual caption, suppressing other audio–visual associations even when the sound and visuals themselves agree.

In contrast, we consider any semantic correspondence between the audio and visual streams, and use its estimated strength to determine how much each pair contributes to learning an audio–visual representation.

\textbf{Learning beyond shared information.} Several recent works aim to go beyond redundant information shared across modalities and also capture unique and/or synergistic information.

FactorCL~\citep{FactorCL} factorizes task-relevant multimodal information into shared and modality-specific components. It uses multimodal augmentations to approximate which information is relevant to the task, and contrastive mutual-information objectives to preserve this information while suppressing task-irrelevant content.
CoMM \citep{CoMM} contrasts augmented fused representations rather than aligning unimodal embeddings pairwise, allowing it to capture synergistic interactions alongside shared and unique information.
InfMasking \citep{infMasking} extends this approach by masking different subsets of modality features during fusion to encourage a broader range of synergistic interactions.
All three methods were intended for classification tasks where their assumptions behind augmentations, such as minimal label preserving assumptions, hold since the modalities of each sample share a single label. We experimentally show that this does not hold in audio-video retrieval, where fine-grained multimodal alignment is important. Instead, we incorporate a prior signal of the degree of agreement between modalities using their captions as a proxy to inform the model on how semantically aligned they are.

\begin{figure}[t]
    \centering
    \vspace{-1em}
    \includegraphics[width=0.95\textwidth]{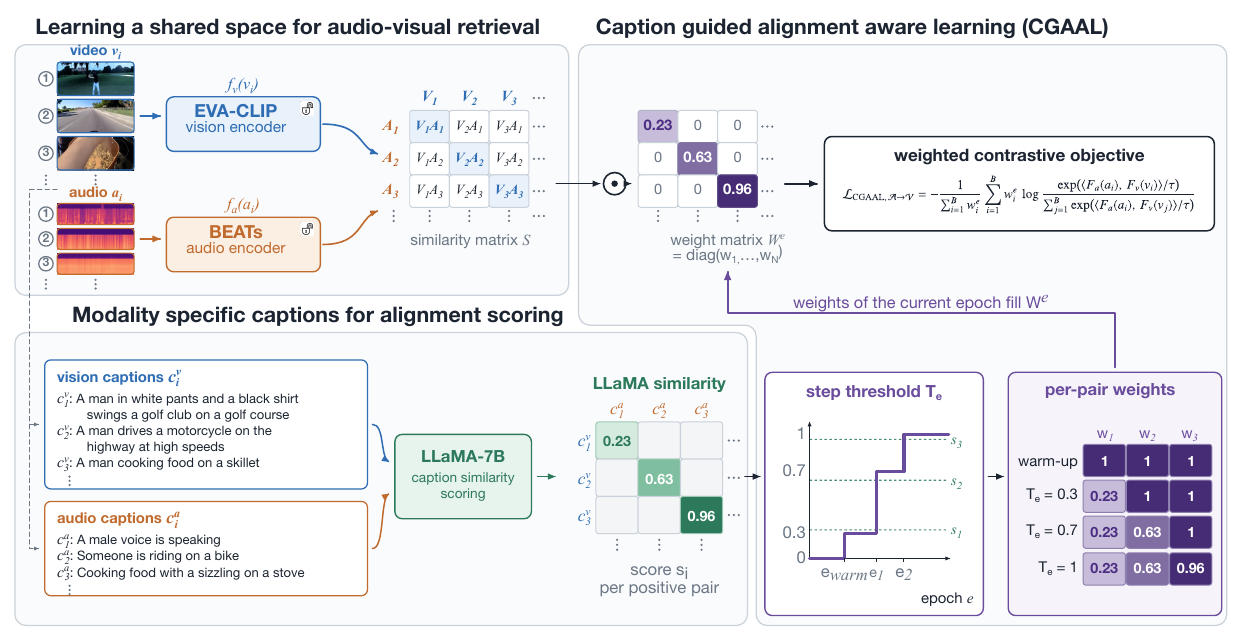}
    \vspace{-0.8em}
    \caption{\textbf{Overview of our caption-guided alignment-aware learning (CGAAL)}. As in standard contrastive learning, we (a) learn a shared space for audio-visual retrieval where audio and video encoders project each modality into a shared embedding space where we measure their similarity. Rather than treating all co-occurring pairs as equally positive, we instead use (b) modality specific captions to estimate the alignment of each positive pair. After first warming up the representation with all positive pairs, CGAAL uses the estimated alignment scores to weight positive pairs in a contrastive loss (c). A gradually increasing threshold introduces the alignment scores gradually so that the weakest correspondences are down-weighted first. This produces a shared embedding space in which proximity better reflects audio–visual correspondence, rather than co-occurrence alone.
    }
    \label{model}
    \vspace{-0.6em}
\end{figure}

\vspace{-0.8em}
\section{Method}\label{method}
\vspace{-0.6em}

Our aim is to learn audio-visual representations that account for varying semantic correspondence between naturally paired modalities, rather than treating all pairs as equally valid positives.
Figure~\ref{model} provides an overview of our approach: given the  audio-visual retrieval setting introduced in Section \ref{task}, we use modality specific captions to estimate the semantic alignment of each pair (Section \ref{captions}) and incorporate these scores into our alignment-aware loss (Section \ref{loss}).

\vspace{-0.4em}
\subsection{Learning a Shared Space for Audio-Visual Retrieval}\label{task}
\vspace{-0.5em}

Given a dataset of $N$ paired audio-visual samples $\{(v_i, a_i)\}_{i=1}^N$, where $v_i$ is a visual frame sampled from a video clip and $a_i$ is the accompanying audio from the same clip, audio-visual retrieval aims to learn a
shared embedding space in which related samples can be found across modalities. To learn this space, modality specific encoders, $F_a$ and $F_v$, are trained to project the audio and vision modalities to features $F_a(a_i)$ and $F_v(v_i)$ in a shared space where similarity can be measured directly. Standard contrastive learning approaches, such as CLIP, do this via an InfoNCE loss. For each audio sample $a_i$ in a batch of size $B$, the visual sample from the same video clip $v_i$ is treated as a positive, while visual samples from other clips $v_j$ serve as negatives. The audio-to-visual loss encourages the positive pair to have higher similarity than these alternatives:
\begin{equation}
    \mathcal{L}^{\text{InfoNCE}}_{\mathcal{A}\to\mathcal{V}} =
-\frac{1}{B} \sum_{i=1}^B
\log \frac{\exp\!\big(\langle F_a(a_i), F_v(v_i)\rangle/\tau\big)}
{\sum_{j=1}^{B}\exp\!\big(\langle F_a(a_i), F_v(v_j)\rangle/\tau\big)}
\end{equation}
Here, $\tau$ is the temperature and $\langle\cdot,\cdot\rangle$ denotes the dot product. The visual-to-audio loss ($\mathcal{L}_{\mathcal{V}\to\mathcal{A}}^\text{InfoNCE}$) is defined analogously and the overall contrastive loss averages these two directions. At inference, this allows for retrieval based on the dot product between representations, ranking candidates according to their similarity to (or distance from) the query.

By using co-occurrence as supervision, InfoNCE learns a shared retrieval space without requiring manual labels. However, the co-occurrence assumption is only appropriate when modalities describe the same semantics, for example, footage of a dog barking corresponds closely to the sounds of its bark.
However, a paired sample may only be partially relevant semantically,
such as a video of someone walking through a park which contains the sound of footsteps alongside a dog barking off-screen.
While it's likely to hear a dog barking in a park, the alignment is only partial as the dog is not visible in view.

Both pairs nevertheless enter InfoNCE as positives when using the co-occurrence supervision, creating pressure to align them to the same degree regardless of how tightly they actually correspond. In this paper, we aim to learn the shared audio-visual embedding space while accounting for this variation in the semantic alignment. This is beneficial for retrieval as candidates are ranked relative to one another. Pulling weakly corresponding pairs together can blur the differences in similarity needed to place a closely matching candidate above a less relevant one.

\vspace{-0.4em}
\subsection{Modality Specific Captions for Alignment Scoring}\label{captions}
\vspace{-0.5em}

To account for varying semantic alignment between paired modalities, we first estimate the degree of semantic correspondence for each audio-visual pair. This can then be used to control how strongly the pair is pulled together during training.
One option is to annotate correspondence manually, however this is expensive. Alternatively, we could use pretrained embeddings, though this estimate could be misleading as unimodal models do not consider cross-modal correspondence, and typical contrastive multimodal models may have already suppressed information unique to each modality.
Instead, we propose to use modality specific captions for the video, $c^v_i$, and audio, $c^a_i$, as proxies of their semantic content.
These captions may be human annotated or generated using modality-specific captioning models.
We hypothesise that unimodal captioning methods have been sufficiently developed to well describe the semantics in a single modality, making these separate captions a practical basis for comparing the semantics of two modalities.
Large-language models (LLMs) can then be used to compare these captions and estimate their potential alignment.

For each audio-visual sample, the LLM is instructed to predict  an alignment score, $s_i$,  in the range $[0,1]$, where 0 indicates no semantic correspondence and 1 indicates complete semantic correspondence.
We leverage in-context learning, providing one highly aligned, one moderately aligned and one weakly aligned caption pair, together with our proposed alignment scores. Figure \ref{llm_scoring_fig} illustrates the LLM-based alignment scoring procedure which we formalise as $F_L$:

\definecolor{pframe}{HTML}{34495E}
\definecolor{pback} {HTML}{F7F8FA}
\definecolor{pstepc}{HTML}{1F4E79}
\definecolor{pvis}  {HTML}{0B6E4F}
\definecolor{paud}  {HTML}{8B3A62}
\definecolor{pscore}{HTML}{B00020}

\newtcolorbox{promptbox}[1][]{%
  enhanced,
  colback=pback, colframe=pframe,
  boxrule=0.6pt, arc=2pt,
  left=5pt, right=5pt, top=4pt, bottom=4pt,
  fonttitle=\bfseries\footnotesize, coltitle=white,
  attach boxed title to top left={xshift=6pt, yshift=-2pt},
  boxed title style={colback=pframe, arc=1pt, boxrule=0pt},
  #1}

\newcommand{\pstep}[2]{%
  \textcolor{pstepc}{\bfseries #1}\,\textcolor{pstepc}{\bfseries #2}\par
  \vspace{-1pt}\textcolor{pframe!40}{\hrule height 0.4pt}\vspace{3pt}}
\newcommand{\plabel}[2]{\textcolor{#1}{\bfseries #2}}
\newcommand{\pscorebox}[1]{%
  \tcbox[on line, size=fbox, colback=pscore!7, colframe=pscore,
         boxrule=0.5pt, arc=1.5pt, left=3pt, right=3pt, top=1pt, bottom=1pt]
        {\bfseries\textcolor{pscore}{#1}}}

\begin{figure}[t]
\vspace{-1em}
\begin{promptbox}[title={Prompt for LLM-based audio--visual alignment scoring}]
\ttfamily\scriptsize\setlength{\parindent}{0pt}\raggedright
\setlength{\emergencystretch}{3em}
\pstep{Step 1:}{Task definition}
Your task is to compare a vision caption and an audio caption and determine
how semantically aligned they are.\par
\vspace{2pt}
Scoring rules:
\begin{itemize}[leftmargin=1.1em, itemsep=0pt, topsep=1pt, parsep=0pt]
  \item Return a decimal number between \texttt{0.0} and \texttt{1.0}.
  \item Consider actions, objects, environments, events, and overall scene meaning.
\end{itemize}
\vspace{4pt}

\pstep{Step 2:}{In-context examples of varying alignment levels}
\plabel{pvis}{$c^v_i$:} A dog runs across a grassy park chasing a ball.\par
\plabel{paud}{$c^a_i$:} A dog barking and running outdoors while people play.\par
\plabel{pframe}{Explanation:} Both captions describe a dog outdoors engaged in
active play in a park.\par
\plabel{pframe}{Score:} 0.91\par
\vspace{4pt}

\pstep{Step 3:}{Inference on a vision--audio caption pair from the dataset}
\plabel{pvis}{$c^v_i$:} A man in white pants and a black shirt swings a
golf club on a golf course.\par
\plabel{paud}{$c^a_i$:} A male voice is speaking.\par
\plabel{pframe}{Score:} \rule{1.2em}{0.4pt}\par
\vspace{5pt}

\noindent\textcolor{pframe!40}{\rule{\linewidth}{0.4pt}}\par
\vspace{4pt}
\normalfont\footnotesize
\textbf{LLM output:}\quad \pscorebox{0.23}
\end{promptbox}
\vspace{-1.4em}
\caption{\textbf{Alignment score estimation.} We prompt the LLM to predict the alignment score by:
(1) setting the criteria for measuring the semantic alignment with clear inputs (modality captions) and the output (alignment score); (2) providing examples with different alignment strength alongside an explanation to calibrate the scores across various examples; and (3) providing the vision and audio caption pair for which a score needs to be predicted.}
\label{llm_scoring_fig}
\end{figure}

\vspace{-1em}
\begin{equation}
s_i = F_L(c^v_i, c^a_i) \in [0,1].
\label{eq:llm_scoring}
\end{equation}
\vspace{-1em}

\begin{figure}[t]
    \centering
    \vspace{-1em}
    \includegraphics[width=0.9\textwidth]{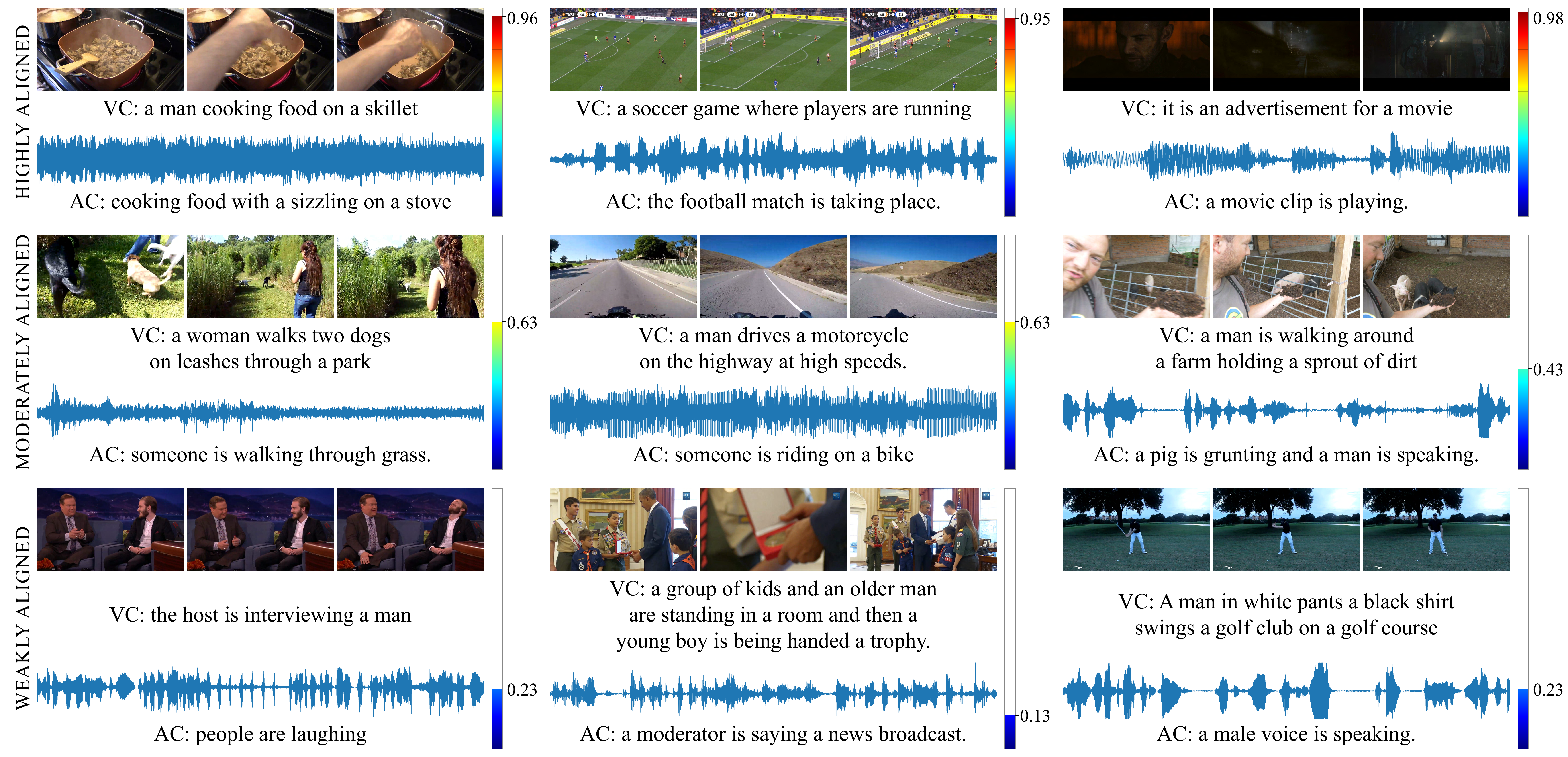}
    \vspace{-1em}
    \caption{\textbf{Examples of modality-specific captions with varying semantic alignment.}
    We show high (top), moderate (middle) and weak (bottom) alignment examples.
    High (row1,col1): vision shows cooking on stove while audio has distinct sizzling sound.
    Moderate (row2,col3): walking is not evident from the audio.
    Weak (Row3 col 3): no evidence of golfing in the audio.}
    \vspace{-0.5em}

    \label{samples_with_scores}
\end{figure}

We show sample prediction scores in Figure \ref{samples_with_scores} which illustrates how the modality-specific captions act as proxies for the semantic content of the underlying vision and audio modalities, and how differences in their semantic correspondence are reflected in the resulting alignment scores. Highly aligned pairs contain substantial shared semantic information across the captions, while partially aligned pairs contain both shared and modality-specific information. For example, in the moderately aligned sample in the second column, the vision caption uniquely provides the highway context, which cannot be heard in the audio, while the remaining information is shared across modalities. In weakly aligned pairs, the captions contain less shared semantic information, indicating lower estimated semantic correspondence between the underlying modalities.

\vspace{-0.5em}
\subsection{Caption Guided Alignment Aware Learning (CGAAL)}\label{loss}
\vspace{-0.5em}

We use our caption-guided alignment scores to vary the contribution of positive pairs to the contrastive objective: pairs with less shared content provide weaker evidence for learning cross-modal correspondence. We first define how the scores are used to weight the loss function, then describe how an initial warm-up and gradual thresholding  introduce these weights during training.

\noindent\textbf{Alignment weighting.} Each positive pair provides a different amount of evidence for audio–visual correspondence. We therefore weight each pair’s contribution to the contrastive loss according to its caption-derived alignment score. Let $w_i$ denote  the weight of the modality pairs,  $(v_i, a_i)$, of the positive sample in the batch $B$. Then the audio-to-video loss becomes:

\vspace{-2em}
\begin{align}
\mathcal{L}_{\text{CGAAL},\,\mathcal{A}\to\mathcal{V}} &=
-\frac{1}{\sum_{i=1}^{B} w_i}
\sum_{i=1}^{B} w_i \,
\log \frac{\exp\!\big(\langle F_a(a_i), F_v(v_i)\rangle/\tau\big)}
{\sum_{j=1}^{B}\exp\!\big(\langle F_a(a_i), F_v(v_j)\rangle/\tau\big)}
\label{eq:cgaal}
\end{align}
\vspace{-1em}

Again $\tau$ is the temperature, the video-to-audio loss is defined analogously and the overall loss is the mean of audio-to-video and video-to-audio losses. Due to the simplicity of our reweighting approach, it can be easily integrated with different  contrastive learning objectives.
In the simplest case our alignment score is used as a pair's weight, i.e. $w_i=s_i$ (Equation~\ref{eq:llm_scoring}).
Next we describe how this varies under gradual thresholding and warm-up.

\textbf{Warm-up period.} At the start of training, the encoders have not yet learned a shared audio–visual space.
We thus train a model for $e_{\mathrm{warmup}}$ epochs with standard InfoNCE, setting $w_i = 1$ for all positive pairs during this short period. This warms up the encoders with the training data, removing any modality shifts between pretraining and training data.

\textbf{Gradual thresholding.}
After the warm-up period, we progressively introduce the alignment scores $s_i$ (Equation~\ref{eq:llm_scoring}) as sample weights. This allows us to progressively relax the assumption of full correspondence, aligning each pair according to how much semantic content its modalities share. We initialize an epoch-dependent threshold $T_e$, that determines which samples are weighted by their alignment score. A sample uses its alignment score $s_i$ as its weight $w_i$ only when $s_i < T_e$.
The threshold is initialised to $T_{e_{\text{warmup}}}$ and increased every $\Delta e$ epochs according to $T_e = \min(1, T_{e-\Delta e} + \Delta T)$, where $\Delta T$ is a hyperparameter controlling the rate of increase. The resulting epoch-dependent sample weights are defined as:
\begin{equation}
w_i^e =
\begin{cases}
1, & \text{if } e < e_{\text{warmup}} \text{ or } s_i \geq T_e, \\
s_i, & \text{otherwise}.
\label{eq:weight_epoch}
\end{cases}
\end{equation}

This means that initially only the most poorly aligned samples are down-weighted, while the remaining samples contribute fully to training. As the threshold increases, higher-alignment samples receive their alignment scores as weights until $T_e=1$. Since $w_i$ varies across epochs, we introduce epoch-dependent weights $w_i^e$ into our loss:

\vspace{-2em}
\begin{align}
\mathcal{L}_{\text{CGAAL},\,\mathcal{A}\to\mathcal{V}} &=
-\frac{1}{\sum_{i=1}^{B} w_i^e}
\sum_{i=1}^{B} w_i^e \,
\log \frac{\exp\!\big(\langle F_a(a_i), F_v(v_i)\rangle/\tau\big)}
{\sum_{j=1}^{B}\exp\!\big(\langle F_a(a_i), F_v(v_j)\rangle/\tau\big)}
\label{eq:cgaal_epoch}
\end{align}

Overall,  this  method encourages the model to focus on bringing the audio and vision representations of aligned samples closer together in the embedding space since they have a higher weight in the loss. When the modalities contain more unique information, the alignment score is lower, and the loss places less emphasis on pulling that pair together. In this way, CGAAL reduces the pressure to align all co-occurring pairs as though they are only multiple views of the same shared information.

\vspace{-0.5em}
\section{Experiments}
\vspace{-0.6em}

We first describe the dataset used and implementation details before evaluating CGAAL against contrastive learning baselines and analysing its performance. We then assess its zero-shot transfer ability and ablate the main components of CGAAL.
\vspace{-0.5em}
\subsection{Dataset Details}
\vspace{-0.5em}
\noindent \textbf{Pretraining.} We use a subset of the VAST-27M dataset \citep{VAST} constituting a randomly selected set of 100K training samples and 1927 test samples. The dataset provides 5 vision captions and 5 audio captions per sample which were automatically generated using modality-specific captioning models. We compute an alignment score for each of the 25 possible audio-visual caption pairs and average these scores to obtain a single alignment score for each sample.

\noindent \textbf{Synthetic Misalignment.} We create a second pretraining set to stress test our method's ability to learn a robust embedding space when challenged with additional misaligned samples. For 15\% of the training samples, we randomly shuffle the audio among samples.
We chose 15\% to resemble the expected noise level in data collected from the internet  \citep{NEURIPS2021_f5e62af8, sharma-etal-2018-conceptual}.

This gives us two training conditions: (1) Without Synthetic Misalignment, which only includes the natural misalignment in the VAST-27M subset and (2) With Synthetic Misalignment, which includes an additional synthetic source of misalignment for 15\% of the training data. In both settings, evaluation is performed on the original, uncorrupted test set.

\begin{figure}[t]
    \centering
    \vspace{-1em}
    \begin{subfigure}[t]{0.48\textwidth}
        \centering
        \includegraphics[width=\textwidth]{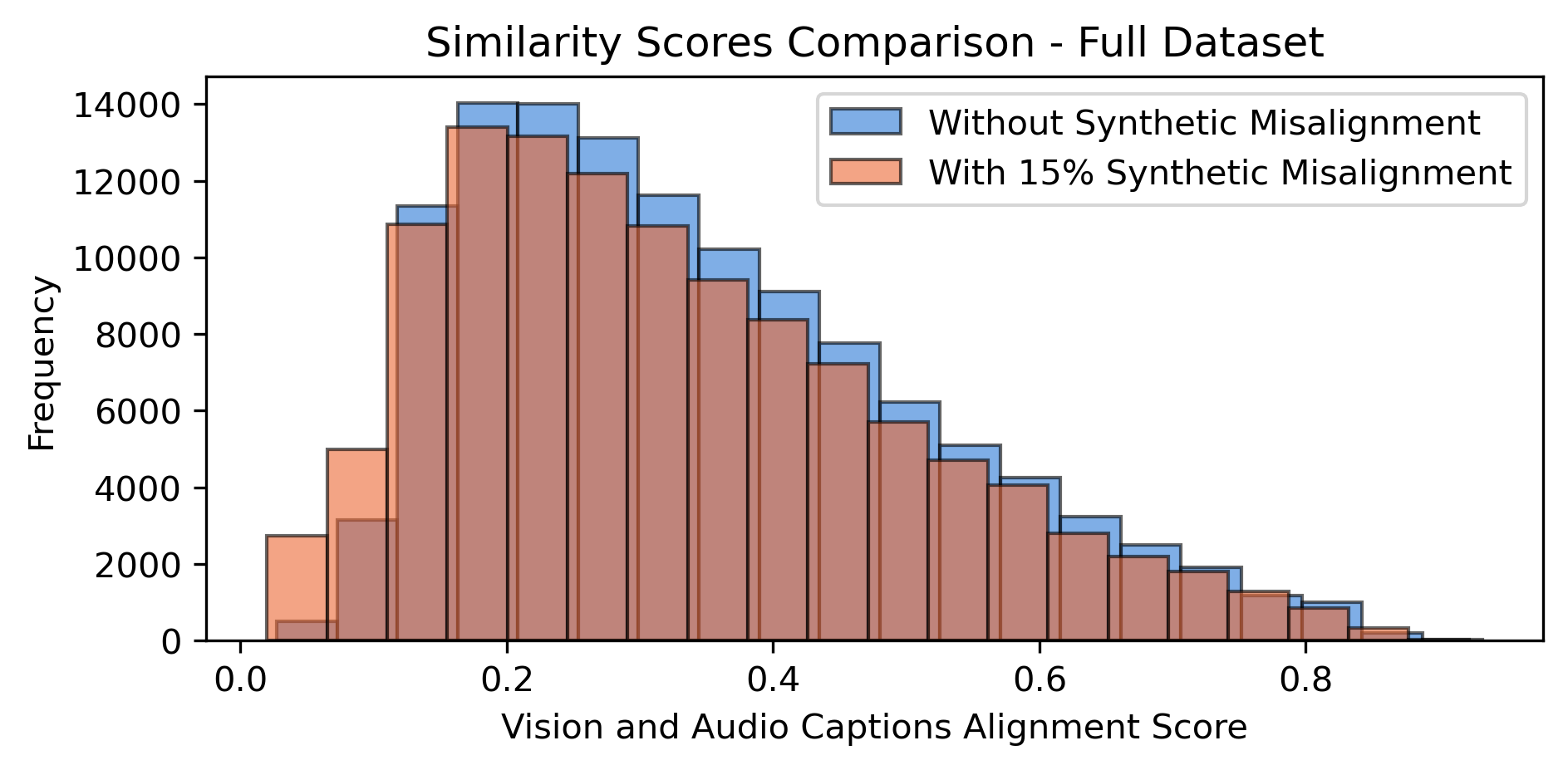}
        \label{full_dataset_dist}
    \end{subfigure}
    \hfill
    \begin{subfigure}[t]{0.48\textwidth}
        \centering
        \includegraphics[width=\textwidth]{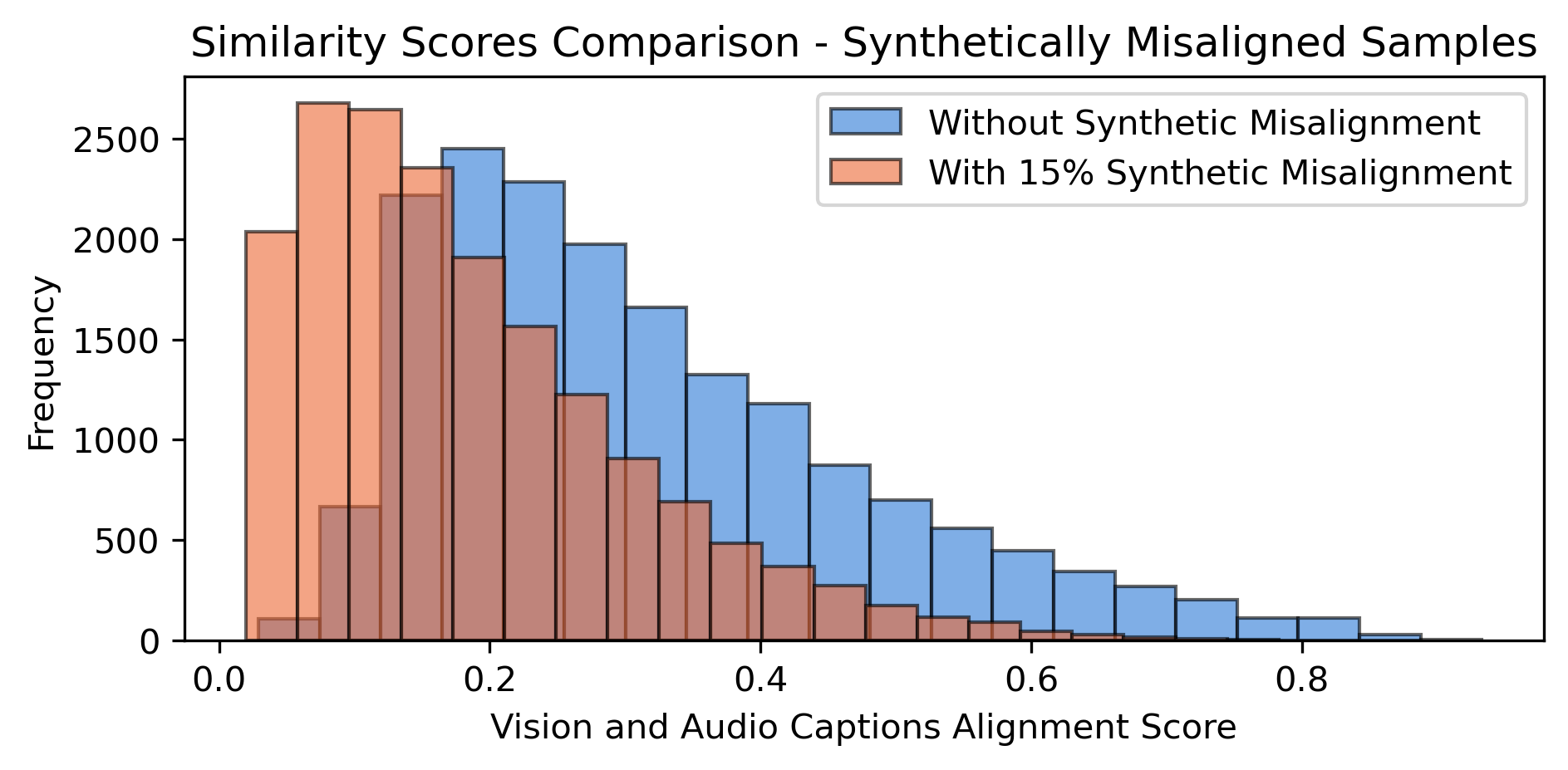}
        \label{misaligned_dist_fig}
    \end{subfigure}
    \vspace{-2em}
    \caption{\textbf{Impact of Synthetic Misalignment on Training Set}. We show the distributions of caption-based alignment scores for the full pretraining set (left) and the 15\% of the samples selected for audio shuffling (right). The original pretraining set (blue) already contains many low-scoring pairs, suggesting substantial natural variation in audio–visual correspondence. Adding synthetic misalignment (orange) shows a clear shift towards lower scores. }
    \label{fig:combined}
    \vspace{-1.5em}
\end{figure}

Figure \ref{fig:combined} compares the distribution of  alignment scores for the two pretraining sets. Without synthetic misalignment the pretraining set already contains many pairs with low estimated alignment, motivating a method that accounts for varying correspondence rather than treating every pair identically. Shuffling the audio shifts the selected pairs further towards lower scores, creating a more challenging scenario.

\noindent \textbf{Zero-shot.} We test whether CGAAL improves audio-video retrieval on datasets unseen during training with  AVE \citep{ave}, DiDeMo \citep{didemo} and MSR-VTT \citep{msrvtt}. AVE focuses on 28 types of audio events which are also clearly visible. DiDeMo contains unedited personal videos from Flickr while MSR-VTT contains short web videos spanning diverse categories.  After excluding clips with no audible audio, the evaluation sets contain 402, 997 and 2617 videos for AVE, DiDeMo and MSR-VTT respectively.

\vspace{-0.3em}
\subsection{Experimental Setup}
\vspace{-0.5em}

\noindent \textbf{Implementation Details.} Following VAST \citep{VAST}, we use BEATs as the audio encoder $F_a$ and the EVA-CLIP vision encoder as $F_v$ with their original pretrained weights. We fine-tune both encoders end-to-end during pretraining for 48 epochs using Adam with a batch size of 8, a learning rate of 5e-6 and weight decay of 1e-2. For each model we integrate our method with, we tune the hyperparameters: $e_{\mathrm{warmup}}$, $T_{e_{\text{warmup}}}$, $\Delta e$ and $\Delta T$. Unless otherwise specified, we use our model in combination with a CLIP-like contrastive objective where we set $e_{\text{warmup}} = 2$, $T_{e_{\text{warmup}}} = 0.4$, $\Delta e = 2$ and $\Delta T = 0.3$. For SLIP (+Ours) and CMC (+Ours), we use  $e_{\mathrm{warmup}}${=}2, $T_{e_{\text{warmup}}} = 0.2$,  $\Delta e${=}2 and  $\Delta T${=}0.2. We choose LLaMA-7B as the LLM for the alignment scores prediction.

\noindent \textbf{Evaluation Metrics.} We report recall R@K for K=1,5,10 alongside median rank for both audio-to-video and video-to-audio retrieval. To summarize recall, we use RSum: the sum of all six recall values across both directions.

\vspace{-0.5em}
\subsection{Comparative Results}
\vspace{-0.5em}

\textbf{Baselines.} We compare CGAAL with three groups of methods.
CoMM~\citep{CoMM} and InfMasking~\citep{infMasking} aim to learn beyond what modalities share, additionally aiming to capture unique and synergistic information in the learned embeddings. They provide a comparison with approaches that also question whether shared information alone is sufficient for multimodal learning, but pursue this through different objectives. We additionally compare to VDR \citep{https://doi.org/10.48550/arxiv.2212.07699}, which learns disentangled semantic representations for cross-modal retrieval. This comparison examines whether semantic disentanglement can address variable cross-modal correspondence without explicitly modelling how strongly each pair should be aligned.
Our standard contrastive baselines are Cross+Self \citep{cross_self}, SLIP \citep{SLIP}, CMC \citep{CMC} and CLIP \citep{CLIP} following~\citet{CoMM} and \citet{infMasking}.  These methods differ in which views they contrast, how the negatives are formed and whether cross-modal learning is combined with within-modal objectives.
CLIP uses one cross-modal term with a learned temperature and negatives only from the opposing modality. CMC averages a fixed-temperature term over all view pairs, with negatives from both views. SLIP adds a separately-scaled SimCLR term to CLIP.
Cross+Self uses one contrastive term throughout, pairing a cross-modal term with an averaged intra-modal term for each modality.
We apply CGAAL to CLIP, SLIP, and CMC to examine whether accounting for pairwise correspondence benefits different contrastive objectives. All methods use the same pretrained EVA-CLIP vision encoder and BEATs audio encoder for a fair comparison.

\begin{table*}[t]
\centering
\setlength{\tabcolsep}{3.5pt}
\renewcommand{\arraystretch}{0.7}
\vspace{-0.5em}
\caption{\textbf{Comparative Results.} Combining our approach with standard contrastive baselines (SLIP, CMC and CLIP) improves retrieval performance both with and without synthetic misalignment.
}
\vspace{-0.7em}
\resizebox{\textwidth}{!}{%

\begin{tabular}{@{}l *{18}{c}@{}}
\toprule

&
\multicolumn{9}{c}{\textbf{Without Synthetic Misalignment}}
&
\multicolumn{9}{c}{\textbf{With Synthetic Misalignment}}
\\
\cmidrule(lr){2-10}
\cmidrule(lr){11-19}

&
\multicolumn{4}{c}{\textbf{A2V}}
&
\multicolumn{4}{c}{\textbf{V2A}}
&
&
\multicolumn{4}{c}{\textbf{A2V}}
&
\multicolumn{4}{c}{\textbf{V2A}}
&
\\
\cmidrule(lr){2-5}
\cmidrule(lr){6-9}
\cmidrule(lr){11-14}
\cmidrule(lr){15-18}

&
R1 $\uparrow$ & R5 $\uparrow$ & R10 $\uparrow$ & MR $\downarrow$
&
R1 $\uparrow$ & R5 $\uparrow$ & R10 $\uparrow$ & MR $\downarrow$
&
\textbf{RSum} $\uparrow$
&
R1 $\uparrow$ & R5 $\uparrow$ & R10 $\uparrow$ & MR $\downarrow$
&
R1 $\uparrow$ & R5 $\uparrow$ & R10 $\uparrow$ & MR $\downarrow$
&
\textbf{RSum} $\uparrow$
\\
\midrule

\textbf{InfMasking}
& 1.2 & 5.7 & 9.0 &159 & 1.9 & 6.18 & 10.4 & 153 & 34.4 &  1.1 & 4.1 &6.6 & 233& 1.3 & 4.4 &7.1 &253 & 24.6 \\

\textbf{CoMM}
& 5.7 & 18.3 & 26.5 & 41 & 5.2 & 17.7 & 26.3 & 38 & 99.7 & 4.3 &14.3 &22.6 & 51& 5.1 & 16.0 &24.4 & 47 & 86.7 \\

\textbf{Cross+Self}
& 10.5 & 28.2 & 39.2 &18 & 10.3 & 28.4 &39.3 &19 & 155.9 &10.0 &27.0 &38.0 & 21& 9.0 & 26.2 &37.1 &21 & 147.3 \\

\textbf{VDR}
& 13.1 & 33.4 & 46.7 &12& 12.5 & 33.6 & 46.0 & 13 & 185.3 & 12.6 & 32.0 & 43.0 & 18 & 11.5 & 30.7 & 42.4 & 17 & 172.2 \\
\midrule

\textbf{SLIP}
& 13.7 & 36.1 & 47.6 & 13 & 14.5 & 35.8 & 47.5 & 12 & 195.2 & 11.7 & 30.4 & 41.6 & 17 & 11.9 & 30.9 & 40.7 & 18 & 167.2 \\

\textbf{SLIP
(+Ours)}
& 15.4 & 36.9  & 48.5 & 11 &  15.9 & 37.3 & 48.9 & 11 & 202.9 & 13.9 & 32.8 & 44.1  & 15 & 13.8& 33.0 & 44.1 & 15 & 181.7 \\

\color{deltagrey}\textbf{$\Delta$SLIP}
& \color{deltagrey}+1.7 & \color{deltagrey}+0.8 & \color{deltagrey}+0.9 & \color{deltagrey}$+$2 & \color{deltagrey}+1.4 & \color{deltagrey}+1.5 & \color{deltagrey}+1.4 & \color{deltagrey}$+$1 & \color{deltagrey}+7.7 & \color{deltagrey}+2.2 & \color{deltagrey}+2.4 & \color{deltagrey}+2.5 & \color{deltagrey}$+$2 & \color{deltagrey}+1.9 & \color{deltagrey}+2.1 & \color{deltagrey}+3.4 & \color{deltagrey}$+$3 & \color{deltagrey}+14.5 \\
\midrule

\textbf{CMC}
& 14.7 & 37.6 & 49.5 &11 & 16.9 & 38.3 &49.8 &11 & 206.8 &13.0 &34.4 &45.4 & 14& 13.9 & 34.4&46.2 &\textbf{13} & 187.3 \\

\textbf{CMC
(+Ours)}
& 15.1 & 37.1 & 50.2 & 10 & 17.0 & 39.2 & 50.6 & 10 & 209.2 & 13.5 & 34.2 & 46.0 & 13 & 14.1 & 35.1  & 46.5 & \textbf{13} & 189.4 \\

\color{deltagrey}\textbf{$\Delta$CMC}
& \color{deltagrey}+0.4 & \color{deltagrey}$-$0.5 & \color{deltagrey}+0.7 & \color{deltagrey}$+$1 & \color{deltagrey}+0.1 & \color{deltagrey}+0.9 & \color{deltagrey}+0.8 & \color{deltagrey}$+$1 & \color{deltagrey}+2.4 & \color{deltagrey}+0.5 & \color{deltagrey}$-$0.2 & \color{deltagrey}+0.6 & \color{deltagrey}$+$1 & \color{deltagrey}+0.2 & \color{deltagrey}+0.7 & \color{deltagrey}+0.3 & \color{deltagrey}0 & \color{deltagrey}+2.1 \\
\midrule

\textbf{CLIP}
& 15.9 & 38.3 & 50.5 & 10 & 17.2 & 38.8 & 50.8 & 10 & 211.5 & 13.0 & 33.5 & 44.4 & 14 & 13.3 & 33.0 & 43.5 & 16 & 180.7 \\

\textbf{CLIP
(+Ours)}
& \textbf{18.4}  & \textbf{41.1} & \textbf{52.8} & \textbf{9} & \textbf{18.4} & \textbf{41.4} & \textbf{52.0} & \textbf{9} &  \textbf{224.1} & \textbf{14.3} & \textbf{35.9} & \textbf{47.6} &  \textbf{12} & \textbf{14.8} & \textbf{36.0} & \textbf{46.6} & \textbf{13} & \textbf{195.2} \\

\color{deltagrey} $\Delta$CLIP
& \color{deltagrey} +2.5 & \color{deltagrey} +2.8 & \color{deltagrey} +2.3 & \color{deltagrey} +1 & \color{deltagrey} +1.2 & \color{deltagrey} +2.6 & \color{deltagrey} +1.2 & \color{deltagrey} +1 & \color{deltagrey} +12.6 & \color{deltagrey} +1.3 & \color{deltagrey} +2.4 & \color{deltagrey} +3.2 & \color{deltagrey} +2 & \color{deltagrey} +1.5 & \color{deltagrey} +3.0 & \color{deltagrey} +3.1 & \color{deltagrey} +3 & \color{deltagrey} +14.5 \\

\bottomrule
\end{tabular}
}
\vspace{-0.7em}
\label{tab:main_results}
\end{table*}

\noindent\textbf{Results.}
Table \ref{tab:main_results} reports results for audio-to-video (A2V) and video-to-audio (V2A) retrieval on the VAST-27M subset after training with and without synthetic misaligment.
CoMM and InfMasking obtain lower results than standard contrastive baselines both with and without synthetic misalignment. This suggests that their approaches to capturing unique and synergistic information do not transfer well to retrieval. Their objectives focus on improving the combined representation of multiple modalities which does not necessarily help the audio rank its matching video above others. VDR achieves stronger retrieval than CoMM and InfMasking but remains below CMC and CLIP in both settings. This suggests that explicitly disentangling semantic content does not, by itself, provide an advantage over the strongest standard contrastive objectives for audio–visual retrieval.
Adding CGAAL consistently improves the retrieval performance of SLIP, CMC and CLIP. These methods differ in how they construct their contrastive objectives, so the gains indicate that accounting for alignment is useful across several ways of learning audio-visual representations. CLIP is the strongest contrastive baseline, leading to CLIP (+Ours) achieving the best overall performance for both V2A and A2V retrieval for all but one metric.

Importantly, CGAAL improves retrieval even when no synthetic misalignment is introduced. Co-occurring audio and vision in the original data do not always describe the same content, so positive pairs already differ in how much evidence they provide for cross-modal alignment. We make this challenge more pronounced by introducing synthetic misalignment. With this, performance falls for the standard contrastive baselines, but adding CGAAL improves both A2V and V2A retrieval. The gains in both settings support the central premise of our method: audio and video should be aligned according to how closely their content matches, rather than simply because they occur together.

\begin{table*}[t]
\centering
\small
\renewcommand{\arraystretch}{0.6}
\setlength{\tabcolsep}{3pt}
\caption{\textbf{Zero-Shot Retrieval.} Adding CGAAL improves a model's ability to generalise to unseen datasets, particularly with increased misalignment in training.}
\vspace{-0.8em}
\resizebox{\textwidth}{!}{%
\begin{tabular}{@{}c l *{18}{c}@{}}
\toprule

& &
\multicolumn{9}{c}{\textbf{Without Synthetic Misalignment}}
&
\multicolumn{9}{c}{\textbf{With Synthetic Misalignment}}
\\
\cmidrule(lr){3-11}
\cmidrule(lr){12-20}

& &
\multicolumn{4}{c}{\textbf{A2V}}
&
\multicolumn{4}{c}{\textbf{V2A}}
&
&
\multicolumn{4}{c}{\textbf{A2V}}
&
\multicolumn{4}{c}{\textbf{V2A}}
&
\\
\cmidrule(lr){3-6}
\cmidrule(lr){7-10}
\cmidrule(lr){12-15}
\cmidrule(lr){16-19}

& &
R1 $\uparrow$ & R5 $\uparrow$ & R10 $\uparrow$ & MR $\downarrow$
&
R1 $\uparrow$ & R5 $\uparrow$ & R10 $\uparrow$ & MR $\downarrow$
&
\textbf{RSum} $\uparrow$
&
R1 $\uparrow$ & R5 $\uparrow$ & R10 $\uparrow$ & MR $\downarrow$
&
R1 $\uparrow$ & R5 $\uparrow$ & R10 $\uparrow$ & MR $\downarrow$
&
\textbf{RSum} $\uparrow$
\\
\midrule

\multirow{3}{*}{\rotatebox[origin=c]{0}{\textbf{AVE}}}
& \textbf{CLIP}
& 5.5 & 21.1 & 31.8 & 25 & \textbf{8.5} & 25.4 & 36.3 & 22 & 128.6 & 5.2 & \textbf{20.1} & 28.6 & 31 & 5.7 & 20.6 & 31.3 & 27 & 111.7 \\
& \textbf{CLIP (+Ours)}
& \textbf{7.2} & \textbf{22.9} & \textbf{34.1} & \textbf{20} & 6.5 & \textbf{26.6} & \textbf{39.5} & \textbf{18} & \textbf{136.8} & \textbf{8.5} & \textbf{20.1} & \textbf{31.1} & \textbf{22} & \textbf{7.0} & \textbf{24.9} & \textbf{35.8} & \textbf{20.5} & \textbf{127.4} \\
& \color{deltagrey}\textbf{$\Delta$CLIP}
& \color{deltagrey}+1.7 & \color{deltagrey}+1.8 & \color{deltagrey}+2.3 & \color{deltagrey}+5 & \color{deltagrey}-2.0 & \color{deltagrey}+1.2 & \color{deltagrey}+3.2 & \color{deltagrey}+4 & \color{deltagrey}+8.2 & \color{deltagrey}+3.3 & \color{deltagrey}+0.0 & \color{deltagrey}+2.5 & \color{deltagrey}+9 & \color{deltagrey}+1.3 & \color{deltagrey}+4.3 & \color{deltagrey}+4.5 & \color{deltagrey}+6.5 & \color{deltagrey}+15.7 \\
\midrule

\multirow{3}{*}{\rotatebox[origin=c]{0}{\textbf{DiDeMo}}}
& \textbf{CLIP}
& 3.0 & 10.7 & 15.8 & 66 & \textbf{3.2} & 10.6 & 17.4 & 58 & 60.8 & 2.1 & 7.8 & 13.1 & 86 & 1.2 & 6.6 & 12.5 & 79 & 43.4 \\
& \textbf{CLIP (+Ours)}
& \textbf{3.5} & \textbf{11.9} & \textbf{18.1} & \textbf{59} & 3.0 & \textbf{12.8} & \textbf{20.8} & \textbf{52} & \textbf{70.1} & \textbf{2.9} & \textbf{11.5} & \textbf{18.1} & \textbf{69} & \textbf{3.0} & \textbf{10.3} & \textbf{15.6} & \textbf{63} & \textbf{61.4} \\
& \color{deltagrey}\textbf{$\Delta$CLIP}
& \color{deltagrey}+0.5 & \color{deltagrey}+1.2 & \color{deltagrey}+2.3 & \color{deltagrey}+7 & \color{deltagrey}-0.2 & \color{deltagrey}+2.2 & \color{deltagrey}+3.4 & \color{deltagrey}+6 & \color{deltagrey}+9.3 & \color{deltagrey}+0.8 & \color{deltagrey}+3.7 & \color{deltagrey}+5.0 & \color{deltagrey}+17 & \color{deltagrey}+1.8 & \color{deltagrey}+3.7 & \color{deltagrey}+3.1 & \color{deltagrey}+16 & \color{deltagrey}+18.0 \\
\midrule

\multirow{3}{*}{\rotatebox[origin=c]{0}{\textbf{MSR-VTT}}}
& \textbf{CLIP}
& 3.8 & 13.4 & 20.3 & 82 & 4.8 & 14.5 & 21.8 & 71.5 & 78.7 & 3.8 & 12.2 & 17.4 & 114 & 4.0 & 12.4 & 18.9 & 112 & 68.7 \\
& \textbf{CLIP (+Ours)}
& \textbf{5.5} & \textbf{15.9} & \textbf{23.0} & \textbf{67} & \textbf{5.2} & \textbf{17.2} & \textbf{23.7} & \textbf{68.5} & \textbf{90.5} & \textbf{5.0} & \textbf{13.5} & \textbf{21.6} & \textbf{79} & \textbf{4.7} & \textbf{14.5} & \textbf{21.3} & \textbf{79} & \textbf{80.7} \\
& \color{deltagrey}\textbf{$\Delta$CLIP}
& \color{deltagrey}+1.7 & \color{deltagrey}+2.5 & \color{deltagrey}+2.7 & \color{deltagrey}+15 & \color{deltagrey}+0.4 & \color{deltagrey}+2.7 & \color{deltagrey}+1.9 & \color{deltagrey}+3 & \color{deltagrey}+11.8 & \color{deltagrey}+1.2 & \color{deltagrey}+1.3 & \color{deltagrey}+4.2 & \color{deltagrey}+35 & \color{deltagrey}+0.7 & \color{deltagrey}+2.1 & \color{deltagrey}+2.4 & \color{deltagrey}+33 & \color{deltagrey}+12.0 \\

\bottomrule
\end{tabular}
}
\label{tab:zero_shot}

\end{table*}

\vspace{-0.5em}
\subsection{Zero-Shot Generalisation}
\vspace{-0.5em}
We examine whether CGAAL's retrieval gains generalise to unseen data on AVE, DiDeMo and MSR-VTT without fine-tuning. Table~\ref{tab:zero_shot} compares CLIP and CLIP (+Ours) trained with and without synthetic misalignment. CGAAL improves RSum on all three datasets in both settings, showing that it results in a more transferable representation. The gains are particularly strong when training includes synthetic misalignment and when the test dataset contains less clear audio-visual correspondences (MSR-VTT and DiDeMo).

\vspace{-0.3em}
\subsection{Ablations}
\vspace{-0.5em}

\noindent\textbf{Model Ablation.} We ablate the components of CGAAL in Table~\ref{tab:ablations}. Starting from the CLIP baseline, we progressively add (1) weighting the loss with alignment scores,  (2) introducing alignment scores with gradual thresholding, (3) adding the warm-up component prior to the gradual thresholding.
Weighting positive pairs by their alignment improves every retrieval metric over CLIP. This demonstrates the effectiveness of using modality-specific captions as a proxy for alignment and incorporating this as a prior signal to reduce the weight of less aligned samples.
Gradual thresholding further improves the overall results (RSum) when synthetic  misalignment is present.  Its greater benefit in the synthetic setting is consistent with progressively limiting the influence of the additional weakly aligned pairs.
Adding the warm-up improves the overall performance with and without synthetic misalignment. This suggests that an initial period of equal weighting may help establish shared structure when positive pairs are generally informative.

\begin{table*}[t]
\centering
\small
\setlength{\tabcolsep}{3.5pt}
\caption{\textbf{Model Ablation.} Alignment weighting contributes most to our performance. Gradual thresholding gives a further gain with synthetic misalignment, while warm-up improves without it.}
\vspace{-0.7em}
\resizebox{\textwidth}{!}{%
\begin{tabular}{@{}
>{\raggedright\arraybackslash}p{3.4cm}
*{18}{c}
@{}}
\toprule

&
\multicolumn{9}{c}{\textbf{Without Synthetic Misalignment}}
&
\multicolumn{9}{c}{\textbf{With Synthetic Misalignment}}
\\
\cmidrule(lr){2-10}
\cmidrule(lr){11-19}

&
\multicolumn{4}{c}{\textbf{A2V}}
&
\multicolumn{4}{c}{\textbf{V2A}}
&
&
\multicolumn{4}{c}{\textbf{A2V}}
&
\multicolumn{4}{c}{\textbf{V2A}}
&
\\
\cmidrule(lr){2-5}
\cmidrule(lr){6-9}
\cmidrule(lr){11-14}
\cmidrule(lr){15-18}

&
R1 $\uparrow$ & R5 $\uparrow$ & R10 $\uparrow$ & MR $\downarrow$
&
R1 $\uparrow$ & R5 $\uparrow$ & R10 $\uparrow$ & MR $\downarrow$
&
\textbf{RSum} $\uparrow$
&
R1 $\uparrow$ & R5 $\uparrow$ & R10 $\uparrow$ & MR $\downarrow$
&
R1 $\uparrow$ & R5 $\uparrow$ & R10 $\uparrow$ & MR $\downarrow$
&
\textbf{RSum} $\uparrow$
\\
\midrule

\textbf{CLIP}
& 15.9 & 38.3 & 50.5 & 10 & 17.2 & 38.8 & 50.8 & 10 & 211.5 & 13.0 & 33.5 & 44.4 & 14 & 13.3 & 33.0 & 43.5 & 16 & 180.7 \\

\textbf{+ alignment weighting}
& 18.0 & 40.9 & 52.0 & 9 & 17.6 &40.3& 52.0 & 9 & 220.8 & 14.8 & 35.0 & 46.6 & 13 & 13.7 & 34.3 & 44.8 & 14 & 189.2 \\

\textbf{+ gradual thresholding}
& 18.1 & 40.6 & 52.3 & 9 & 17.2 & 39.9 & 51.6 & 9 & 219.7  &  14.4 & 35.7  & 47.3 & 12 & 15.0 & 34.9 & 46.2 & 13 & 193.5 \\

\textbf{+ warm up}
& 18.4  & 41.1 & 52.8 & 9 & 18.4 & 41.4 & 52.0 & 9 &  224.1 & 14.3 & 35.9 & 47.6 &  12 & 14.8 & 36.0 & 46.6 & 13 & 195.2 \\

\bottomrule
\end{tabular}
}
\vspace{-0.5em}
\label{tab:ablations}
\end{table*}

\begin{table*}[t]
\centering
\setlength{\tabcolsep}{3.5pt}
\renewcommand{\arraystretch}{0.7}
\vspace{-0.5em}
\caption{\textbf{Alignment Weighting vs. Filtering.} Filtering removes training pairs with alignment scores below threshold \(T\). CGAAL improves retrieval in both settings, whereas filtering at either threshold reduces performance relative to CLIP.}
\vspace{-0.7em}
\resizebox{\textwidth}{!}{%

\begin{tabular}{@{}l *{18}{c}@{}}
\toprule

&
\multicolumn{9}{c}{\textbf{Without Synthetic Misalignment}}
&
\multicolumn{9}{c}{\textbf{With Synthetic Misalignment}}
\\
\cmidrule(lr){2-10}
\cmidrule(lr){11-19}

&
\multicolumn{4}{c}{\textbf{A2V}}
&
\multicolumn{4}{c}{\textbf{V2A}}
&
&
\multicolumn{4}{c}{\textbf{A2V}}
&
\multicolumn{4}{c}{\textbf{V2A}}
&
\\
\cmidrule(lr){2-5}
\cmidrule(lr){6-9}
\cmidrule(lr){11-14}
\cmidrule(lr){15-18}

&
R1 $\uparrow$ & R5 $\uparrow$ & R10 $\uparrow$ & MR $\downarrow$
&
R1 $\uparrow$ & R5 $\uparrow$ & R10 $\uparrow$ & MR $\downarrow$
&
\textbf{RSum} $\uparrow$
&
R1 $\uparrow$ & R5 $\uparrow$ & R10 $\uparrow$ & MR $\downarrow$
&
R1 $\uparrow$ & R5 $\uparrow$ & R10 $\uparrow$ & MR $\downarrow$
&
\textbf{RSum} $\uparrow$
\\
\midrule

\textbf{CLIP}
& 15.9 & 38.3 & 50.5 & 10 & 17.2 & 38.8 & 50.8 & 10 & 211.5 & 13.0 & 33.5 & 44.4 & 14 & 13.3 & 33.0 & 43.5 & 16 & 180.7 \\

\textbf{CLIP + Filtering ($T=0.7$)}
& 3.9 & 12.8 & 18.8 & 90 & 4.3 & 12.1 & 17.4 & 88 & 69.2 & 3.7 & 11.2 & 16.6 & 101 & 3.5 & 10.1 & 16.0 & 100 & 61.1\\

\textbf{CLIP + Filtering ($T=0.4$)}
& 11.6 & 29.2 & 39.4 & 20 & 11.9 & 29.4 & 38.9& 20 & 160.4  & 11.2 & 27.9 & 36.7 & 23 & 11.0 & 27.0 & 37.4 & 24 & 151.2 \\

\textbf{CLIP + CGAAL
(Ours)}
& \textbf{18.4}  & \textbf{41.1} & \textbf{52.8} & \textbf{9} & \textbf{18.4} & \textbf{41.4} & \textbf{52.0} & \textbf{9} &  \textbf{224.1} & \textbf{14.3} & \textbf{35.9} & \textbf{47.6} &  \textbf{12} & \textbf{14.8} & \textbf{36.0} & \textbf{46.6} & \textbf{13} & \textbf{195.2} \\

\bottomrule
\end{tabular}
}
\vspace{-0.7em}
\label{tab:filtering}
\end{table*}

\begin{table*}[t]
\centering
\small
\setlength{\tabcolsep}{3.5pt}
\caption{\textbf{Robustness to LLM Choice.}  CGAAL benefits from LLM-based alignment scoring without relying on a particular LLM.}
\vspace{-1em}
\resizebox{\textwidth}{!}{%
\begin{tabular}{@{}
>{\raggedright\arraybackslash}p{3.4cm}
*{18}{c}
@{}}
\toprule

&
\multicolumn{9}{c}{\textbf{Without Synthetic Misalignment}}
&
\multicolumn{9}{c}{\textbf{With Synthetic Misalignment}}
\\
\cmidrule(lr){2-10}
\cmidrule(lr){11-19}

&
\multicolumn{4}{c}{\textbf{A2V}}
&
\multicolumn{4}{c}{\textbf{V2A}}
&
&
\multicolumn{4}{c}{\textbf{A2V}}
&
\multicolumn{4}{c}{\textbf{V2A}}
&
\\
\cmidrule(lr){2-5}
\cmidrule(lr){6-9}
\cmidrule(lr){11-14}
\cmidrule(lr){15-18}

&
R1 $\uparrow$ & R5 $\uparrow$ & R10 $\uparrow$ & MR $\downarrow$
&
R1 $\uparrow$ & R5 $\uparrow$ & R10 $\uparrow$ & MR $\downarrow$
&
\textbf{RSum} $\uparrow$
&
R1 $\uparrow$ & R5 $\uparrow$ & R10 $\uparrow$ & MR $\downarrow$
&
R1 $\uparrow$ & R5 $\uparrow$ & R10 $\uparrow$ & MR $\downarrow$
&
\textbf{RSum} $\uparrow$
\\
\midrule

 \textbf{CLIP}
 & 15.9 & 38.3 & 50.5 & 10 & 17.2 & 38.8 & 50.8 & 10 & 211.5 & 13.0 & 33.5 & 44.4 & 14 & 13.3 & 33.0 & 43.5 & 16 & 180.7 \\

\textbf{CLIP+Ours (BERT)}
&  17.1 & 42.2 & 52.7  & 9 & 17.9 & 41.7 & 52.4 & 9 & 224.0& 13.8 & 34.4 & 45.7 & 14 & 13.8 & 33.8 & 45.0 & 14 & 186.5 \\

\textbf{CLIP+Ours (Gemini)}
& 17.6 & 41.8 & 53.0 & 9 & 17.2 & 40.7 & 52.1 & 9 & 222.4 & 14.6 & 35.6 & 47.2 & 12  & 14.6 & 35.0 & 46.0 & 13 & 193.0  \\

\textbf{CLIP+Ours (LLaMA)}
& 18.4  & 41.1 & 52.8 & 9 & 18.4 & 41.4 & 52.0 & 9 &  224.1 & 14.3 & 35.9 & 47.6 &  12 & 14.8 & 36.0 & 46.6 & 13 & 195.2 \\

\bottomrule
\end{tabular}
}
\vspace{-1em}
\label{tab:llm_abation}
\end{table*}

\noindent\textbf{Weighting vs. Filtering.}
We verify that weighting training pairs by their alignment scores is more effective than filtering out poorly aligned pairs.  We compare CGAAL with removing pairs whose alignment scores fall below fixed thresholds of 0.4 or 0.7. As shown in Table~\ref{tab:filtering}, both filtering variants reduce retrieval performance relative to CLIP, with the stricter threshold causing a larger drop. CGAAL instead improves performance both with and without synthetic misalignment. These results suggest that partially aligned pairs can still provide useful supervision when their contribution reflects their degree of correspondence.

\noindent\textbf{Robustness to LLM Choice.} We assess whether CGAAL’s effectiveness depends on the alignment scorer by comparing LLaMA, Gemini, and similarity scores computed from BERT embeddings of the audio and visual captions. Table~\ref{tab:llm_abation} shows that all three approaches improve RSum over CLIP in both settings. BERT-based scoring performs comparably to the LLMs without synthetic misalignment, but its gains are smaller when synthetic misalignment is introduced. LLaMA and Gemini achieve similar performance in both settings. These results support BERT as a less computationally demanding alternative, while LLM-based scoring provides greater robustness to misalignment with little sensitivity to the choice of LLM.

\vspace{-0.3em}
\subsection{Analysis of Audio-Visual Alignment}
\vspace{-0.5em}

To better understand our alignment scores and their effect on cross-modal learning, we first assess their agreement with human judgments. We then examine how CGAAL shapes the learned embedding space and affects retrieval across alignment levels through quantitative and qualitative analyses.

\begin{wrapfigure}{r}{0.3\columnwidth}
\centering
        \vspace{-1.5em}
        \includegraphics[width=\linewidth]{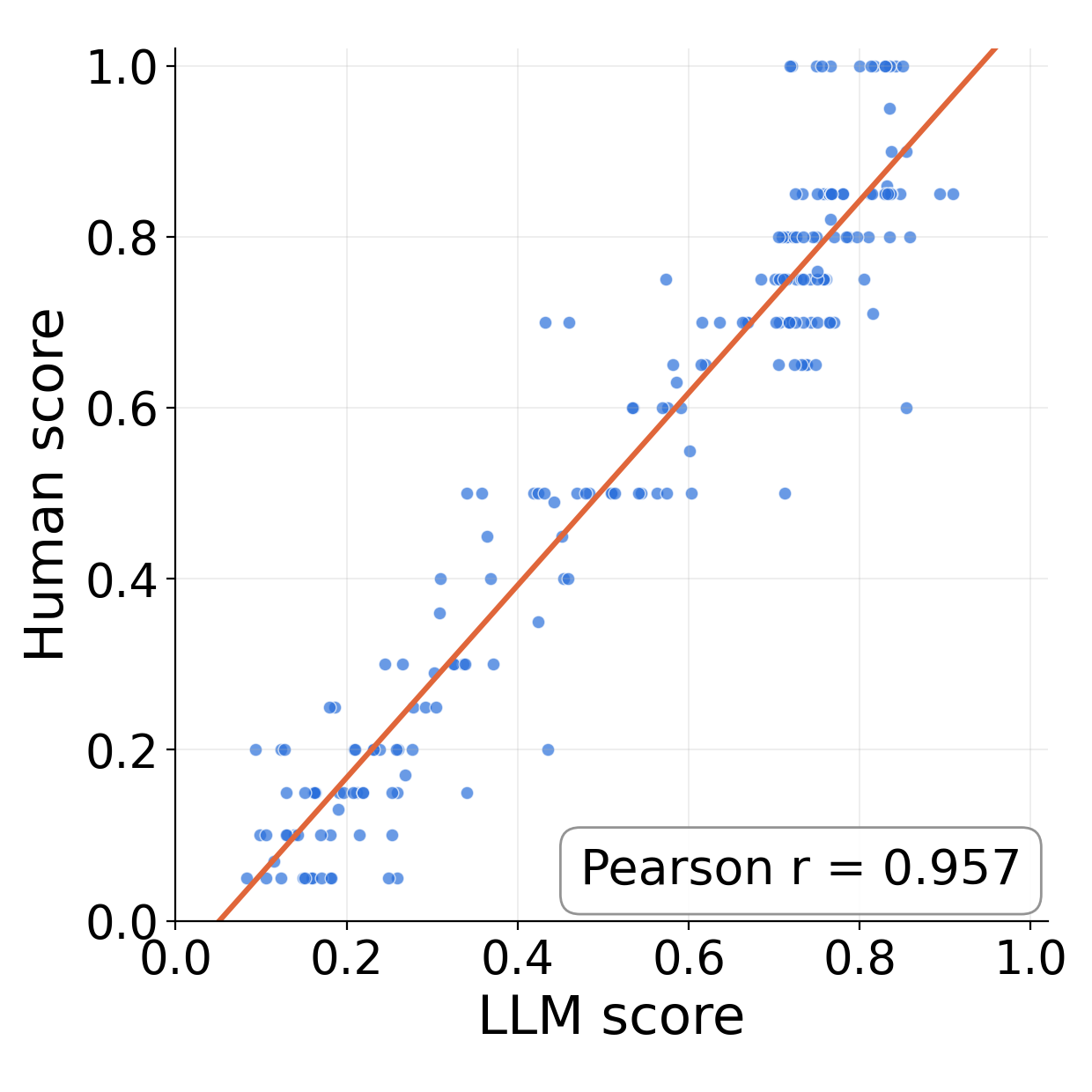}
        \label{human_verification:human_llm_error}
    \vspace{-2em}
    \caption{\textbf{Reliability of Alignment Scores.}
    Caption-based LLM scores closely agree with human ratings of audio–visual correspondence.%
    }
    \vspace{-1em}
    \label{fig:human_verification}
\end{wrapfigure}

\noindent\textbf{Reliability of Alignment Scores}
We assess whether caption-based LLM scores reflect audio-visual correspondence through a verification study on 200 clips. Human evaluators watched each clip with its audio and rated the semantic alignment using a continuous slider. As shown in Figure \ref{fig:human_verification}, the LLM scores correlate strongly with human ratings (Pearson's $r{=}0.957$), with only 11 clips having a score difference greater than 0.2. This agreement supports the use of caption-based scoring to estimate audio–visual alignment.

\begin{wrapfigure}{r}{0.55\columnwidth}
\centering
\vspace{-1em}
    \includegraphics[width=\linewidth]{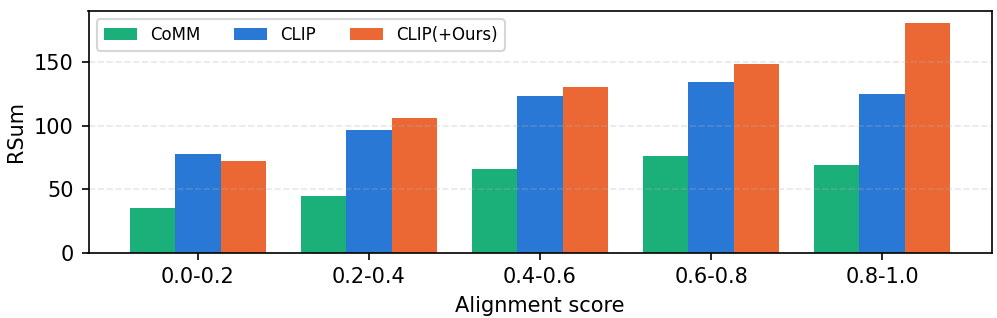}
    \vspace{-1.7em}
    \caption{\textbf{Performance vs. alignment score}. The relative improvement of our method compared to CLIP and CoMM increases with alignment score.}
    \vspace{-0.5em}
    \label{fig:bin_performance}
\end{wrapfigure}

\noindent\textbf{Performance vs. alignment score.} In Figure~\ref{fig:bin_performance} we compare the retrieval performance with RSum for samples with varying levels of alignment score. CLIP has a small advantage for samples with low alignment (0.0--0.2).  These pairs remain correct matches under the retrieval metric despite having little estimated semantic correspondence. Treating every pair equally during training may therefore help CLIP retrieve them. Our approach performs better in every other range, with its largest gain for the most strongly aligned pairs (0.8--1.0). This pattern is consistent with CGAAL placing greater emphasis on pairs whose modalities share more information.

\begin{figure}[t]
    \centering
    \includegraphics[width=\textwidth]{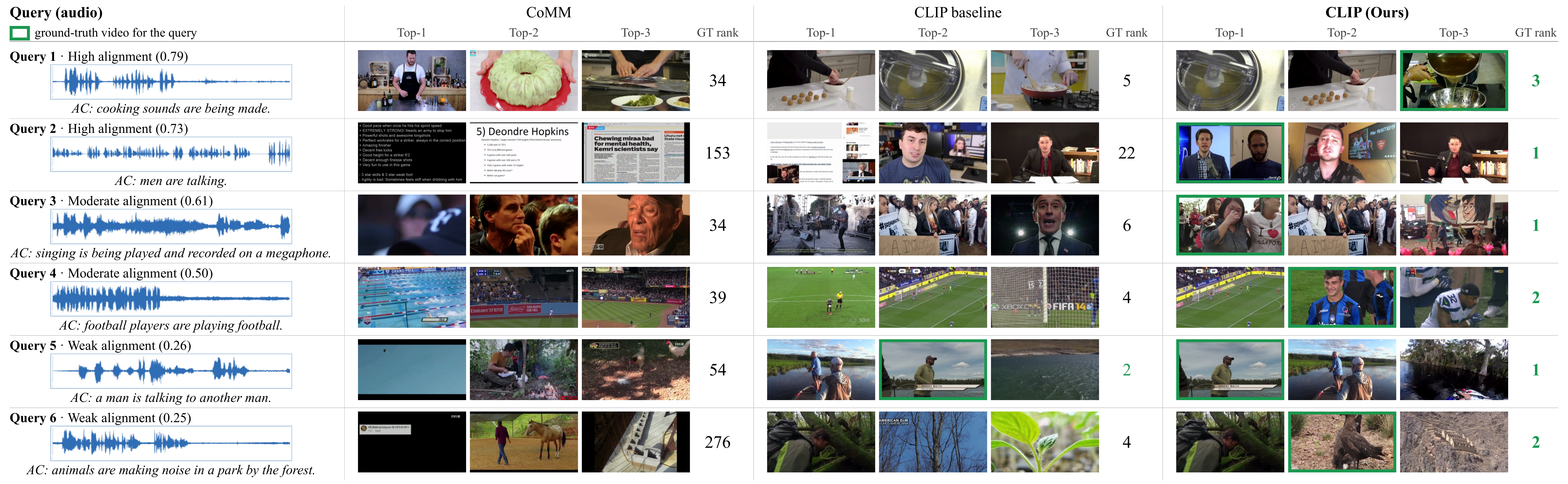}
    \vspace{-2em}
    \caption{\textbf{Qualitative Results} for A2V over different alignment levels.  Our retrieved videos show better fine-grained semantic correspondence with the audio query.
    }
    \label{fig:qual_results}
    \vspace{-1em}

\end{figure}

\noindent\textbf{Qualitative results.} We show audio-to-video retrieval for six samples with different levels of alignment. Audio captions are displayed only to help readers interpret each query. At inference, the model uses only  audio and video modalities, without captions or alignment scores. Across these examples, our approach appears to capture more specific audio–visual correspondence, whereas CLIP favours scenes broadly associated with a sound. For example, given an audio of two men talking, CGAAL ranks a video of two men conversing first, while CLIP retrieves a video showing a webpage accompanied by spoken commentary from one person.

\noindent\textbf{Alignment in the Learned Embedding.}
We compare the cosine similarity of paired audio and video embeddings at different levels of semantic alignment. As shown in Table~\ref{tab:cos_sim_by_alignment_fine}, CGAAL reduces mean similarity across all alignment levels, with a larger relative reduction for weakly aligned pairs (14.4\%) than strongly aligned pairs (6.1\%). This pattern is consistent with reducing the pressure to align modalities that share less semantic content.

\begin{table*}[t]
\centering
\scriptsize
\caption{\textbf{Cross-modal embedding similarity across alignment levels.} Mean ($\pm$ standard deviation) audio--video cosine similarity on test set, grouped by alignment score.
}
\vspace{-1em}
\setlength{\tabcolsep}{6pt}
\begin{tabular}{@{}lccccc@{}}
\toprule
\textbf{Alignment} & \textbf{0.8--1.0} & \textbf{0.6--0.8} & \textbf{0.4--0.6} & \textbf{0.2--0.4} & \textbf{0.0--0.2} \\
\midrule
CLIP          & 0.371 $\pm$ 0.102 & 0.362 $\pm$ 0.124 & 0.344 $\pm$ 0.129 & 0.309 $\pm$ 0.128 & 0.268 $\pm$ 0.132 \\
CLIP (+Ours)  & 0.349 $\pm$ 0.102 & 0.324 $\pm$ 0.111 & 0.304 $\pm$ 0.120 & 0.269 $\pm$ 0.116 & 0.229 $\pm$ 0.114 \\
\midrule
$\Delta$ & $-$0.023 ($-$6.1\%) & $-$0.037 ($-$10.3\%) & $-$0.040 ($-$11.6\%) & $-$0.039 ($-$12.7\%) & $-$0.039 ($-$14.4\%) \\
\bottomrule
\end{tabular}
\label{tab:cos_sim_by_alignment_fine}
\end{table*}

\vspace{-0.5em}
\section{Conclusions}
\vspace{-0.6em}

In this work, we explored the challenge of partial correspondence in multimodal contrastive learning for audio-visual retrieval. We proposed CGAAL, which uses modality-specific captions as proxies to estimate the semantic alignment of the modalities of paired signals and weights pairs in the contrastive loss according to this alignment. This allows for more effective learning of the commonality between modalities, resulting in better audio-video retrieval. Furthermore, we show our method is effective with a variety of contrastive learning approaches, can handle both naturally occurring and additional synthetic misalignment, and improves generalisability to unseen data. Together, these results show the value of accounting for varying degrees of correspondence between positive pairs.

\noindent\textbf{Acknowledgements: }O Emara is supported by UKRI CDT in Interactive AI (EP/\allowbreak S022937/\allowbreak 1). H Doughty is supported by the Dutch Research Council~(NWO) under a Veni grant (VI.Veni.222.160). The authors also acknowledge the use of Isambard-AI National AI Research Resource (AIRR). We acknowledge the usage of GPU hours granted as part of AIRR Innovator project `5D Hand-Object Interaction Modelling from In-the-wild Videos’ and uob-0261-2834-2079-1 `Explainable and Trustworthy Multi-modal Learning'. Isambard-AI is operated by the University of Bristol and is funded by the UK Government's Department for Science, Innovation and Technology (DSIT) via UK Research and Innovation; and the Science and Technology Facilities Council [ST/\allowbreak AIRR/\allowbreak I-A-I/\allowbreak 1023]. We also extend our gratitude to SURF (\url{www.surf.nl}) for granting compute resources from the National Supercomputer Snellius.

\bibliography{iclr2027_conference}

\begin{thebibliography}{38}
\providecommand{\natexlab}[1]{#1}
\providecommand{\url}[1]{\texttt{#1}}
\expandafter\ifx\csname urlstyle\endcsname\relax
  \providecommand{\doi}[1]{doi: #1}\else
  \providecommand{\doi}{doi: \begingroup \urlstyle{rm}\Url}\fi

\bibitem[Alayrac et~al.(2020)Alayrac, Recasens, Schneider, Arandjelovi{\'c}, Ramapuram, De~Fauw, Smaira, Dieleman, and Zisserman]{alayrac2020self}
Jean-Baptiste Alayrac, Adri{\`a} Recasens, Rosalia Schneider, Relja Arandjelovi{\'c}, Jason Ramapuram, Jeffrey De~Fauw, Lucas Smaira, Sander Dieleman, and Andrew Zisserman.
\newblock {S}elf-{S}upervised {M}ulti{M}odal {V}ersatile {N}etworks.
\newblock In \emph{NeurIPS}, 2020.

\bibitem[Araujo et~al.(2025)Araujo, Rouditchenko, Gong, Bhati, Thomas, Kingsbury, Karlinsky, Feris, Glass, and Kuehne]{cav_mae}
Edson Araujo, Andrew Rouditchenko, Yuan Gong, Saurabhchand Bhati, Samuel Thomas, Brian Kingsbury, Leonid Karlinsky, Rogerio Feris, James~R. Glass, and Hilde Kuehne.
\newblock Cav-mae sync: Improving contrastive audio-visual mask autoencoders via fine-grained alignment.
\newblock In \emph{Proceedings of the IEEE/CVF Conference on Computer Vision and Pattern Recognition (CVPR)}, 2025.

\bibitem[Chen et~al.(2021)Chen, Rouditchenko, Duarte, Kuehne, Thomas, Boggust, Panda, Kingsbury, Feris, Harwath, Glass, Picheny, and Chang]{chen2021multimodal}
Brian Chen, Andrew Rouditchenko, Kevin Duarte, Hilde Kuehne, Samuel Thomas, Angie Boggust, Rameswar Panda, Brian Kingsbury, Rogerio Feris, David Harwath, James Glass, Michael Picheny, and Shih-Fu Chang.
\newblock Multimodal clustering networks for self-supervised learning from unlabeled videos.
\newblock In \emph{Proceedings of the IEEE/CVF International Conference on Computer Vision (ICCV)}, pages 8012--8021, October 2021.

\bibitem[Chen et~al.(2024)Chen, Ashutosh, Girdhar, Harwath, and Grauman]{SoundingActions}
Changan Chen, Kumar Ashutosh, Rohit Girdhar, David Harwath, and Kristen Grauman.
\newblock Soundingactions: Learning how actions sound from narrated egocentric videos.
\newblock In \emph{CVPR}, 2024.

\bibitem[Chen et~al.(2023)Chen, Li, Wang, Zhao, Sun, Zhu, and Liu]{VAST}
Sihan Chen, Handong Li, Qunbo Wang, Zijia Zhao, Mingzhen Sun, Xinxin Zhu, and Jing Liu.
\newblock Vast: A vision-audio-subtitle-text omni-modality foundation model and dataset.
\newblock \emph{Advances in Neural Information Processing Systems}, 36:\penalty0 72842--72866, 2023.

\bibitem[Dufumier et~al.(2025)Dufumier, Castillo~Navarro, Tuia, and Thiran]{CoMM}
Benoit Dufumier, Javiera Castillo~Navarro, Devis Tuia, and Jean-Philippe Thiran.
\newblock What to align in multimodal contrastive learning?
\newblock In \emph{International Conference on Learning Representations}, volume 2025, pages 5408--5432, 2025.

\bibitem[Fan et~al.(2026)Fan, Liu, Li, Zhang, Jin, and Zhang]{Fan2026}
Baoyu Fan, Lu~Liu, Xiaochuan Li, Runze Zhang, Liang Jin, and Jin Zhang.
\newblock Fine-grained audio–visual event localization.
\newblock \emph{IEEE Transactions on Neural Networks and Learning Systems}, 37\penalty0 (1):\penalty0 52–66, January 2026.
\newblock ISSN 2162-2388.
\newblock \doi{10.1109/tnnls.2025.3600878}.
\newblock URL \url{http://dx.doi.org/10.1109/TNNLS.2025.3600878}.

\bibitem[Federici et~al.(2020)Federici, Dutta, Forré, Kushman, and Akata]{multiview_red}
Marco Federici, Anjan Dutta, Patrick Forré, Nate Kushman, and Zeynep Akata.
\newblock Learning robust representations via multi-view information bottleneck, 2020.
\newblock URL \url{https://arxiv.org/abs/2002.07017}.

\bibitem[Fradkin et~al.(2024)Fradkin, Azadi, Suri, Wenkel, Bashashati, Sypetkowski, and Beaini]{PhenoMolecular}
Philip Fradkin, Puria Azadi, Karush Suri, Frederik Wenkel, Ali Bashashati, Maciej Sypetkowski, and Dominique Beaini.
\newblock How molecules impact cells: Unlocking contrastive phenomolecular retrieval.
\newblock In A.~Globerson, L.~Mackey, D.~Belgrave, A.~Fan, U.~Paquet, J.~Tomczak, and C.~Zhang, editors, \emph{Advances in Neural Information Processing Systems}, volume~37, pages 110667--110701. Curran Associates, Inc., 2024.
\newblock \doi{10.52202/079017-3513}.
\newblock URL \url{https://proceedings.neurips.cc/paper_files/paper/2024/file/c793577b644268259b1416464a6cdb8c-Paper-Conference.pdf}.

\bibitem[Girdhar et~al.(2023)Girdhar, El-Nouby, Liu, Singh, Alwala, Joulin, and Misra]{ImageBind}
Rohit Girdhar, Alaaeldin El-Nouby, Zhuang Liu, Mannat Singh, Kalyan~Vasudev Alwala, Armand Joulin, and Ishan Misra.
\newblock Imagebind one embedding space to bind them all.
\newblock In \emph{2023 IEEE/CVF Conference on Computer Vision and Pattern Recognition (CVPR)}, pages 15180--15190. IEEE, 2023.

\bibitem[Guzhov et~al.(2022)Guzhov, Raue, Hees, and Dengel]{guzhov2022audioclip}
Andrey Guzhov, Federico Raue, J{\"o}rn Hees, and Andreas Dengel.
\newblock Audioclip: Extending clip to image, text and audio.
\newblock In \emph{ICASSP 2022-2022 IEEE international conference on acoustics, speech and signal processing (ICASSP)}, pages 976--980. IEEE, 2022.

\bibitem[Hendricks et~al.(2018)Hendricks, Wang, Shechtman, Sivic, Darrell, and Russell]{didemo}
Lisa~Anne Hendricks, Oliver Wang, Eli Shechtman, Josef Sivic, Trevor Darrell, and Bryan Russell.
\newblock Localizing moments in video with temporal language.
\newblock In Ellen Riloff, David Chiang, Julia Hockenmaier, and Jun{'}ichi Tsujii, editors, \emph{Proceedings of the 2018 Conference on Empirical Methods in Natural Language Processing}, pages 1380--1390, Brussels, Belgium, October-November 2018. Association for Computational Linguistics.
\newblock \doi{10.18653/v1/D18-1168}.
\newblock URL \url{https://aclanthology.org/D18-1168/}.

\bibitem[Huang et~al.(2021{\natexlab{a}})Huang, Shen, Lungren, and Yeung]{huang2021gloria}
Shih-Cheng Huang, Liyue Shen, Matthew~P Lungren, and Serena Yeung.
\newblock Gloria: A multimodal global-local representation learning framework for label-efficient medical image recognition.
\newblock In \emph{Proceedings of the IEEE/CVF International Conference on Computer Vision}, pages 3942--3951, 2021{\natexlab{a}}.

\bibitem[Huang et~al.(2021{\natexlab{b}})Huang, Niu, Liu, Ding, Xiao, Wu, and Peng]{NEURIPS2021_f5e62af8}
Zhenyu Huang, Guocheng Niu, Xiao Liu, Wenbiao Ding, Xinyan Xiao, Hua Wu, and Xi~Peng.
\newblock Learning with noisy correspondence for cross-modal matching.
\newblock In M.~Ranzato, A.~Beygelzimer, Y.~Dauphin, P.S. Liang, and J.~Wortman Vaughan, editors, \emph{Advances in Neural Information Processing Systems}, volume~34, pages 29406--29419. Curran Associates, Inc., 2021{\natexlab{b}}.
\newblock URL \url{https://proceedings.neurips.cc/paper_files/paper/2021/file/f5e62af885293cf4d511ceef31e61c80-Paper.pdf}.

\bibitem[Li et~al.(2026)Li, Yang, Song, Liu, and Zhang]{text_video}
Guangzhao Li, Yanming Yang, Chenxi Song, Xiaohong Liu, and Chi Zhang.
\newblock Flowdirector: Training-free flow steering for precise text-to-video editing.
\newblock In \emph{Proceedings of the IEEE/CVF Conference on Computer Vision and Pattern Recognition (CVPR)}, pages 7805--7815, June 2026.

\bibitem[Liang et~al.(2023)Liang, Deng, Ma, Zou, Morency, and Salakhutdinov]{FactorCL}
Paul~Pu Liang, Zihao Deng, Martin~Q Ma, James~Y Zou, Louis-Philippe Morency, and Ruslan Salakhutdinov.
\newblock Factorized contrastive learning: Going beyond multi-view redundancy.
\newblock \emph{Advances in Neural Information Processing Systems}, 36:\penalty0 32971--32998, 2023.

\bibitem[Liu et~al.(2024)Liu, Chen, He, Guo, Zhu, Wang, and Tang]{VALOR}
Jing Liu, Sihan Chen, Xingjian He, Longteng Guo, Xinxin Zhu, Weining Wang, and Jinhui Tang.
\newblock Valor: Vision-audio-language omni-perception pretraining model and dataset.
\newblock \emph{IEEE Transactions on Pattern Analysis and Machine Intelligence}, 47\penalty0 (2):\penalty0 708--724, 2024.

\bibitem[Lu et~al.(2025)Lu, Weinberger, Kim, and Lee]{CellCLIP}
MingYu Lu, Ethan Weinberger, Chanwoo Kim, and Su-In Lee.
\newblock Cellclip - learning perturbation effects in cell painting via text-guided contrastive learning.
\newblock In D.~Belgrave, C.~Zhang, H.~Lin, R.~Pascanu, P.~Koniusz, M.~Ghassemi, and N.~Chen, editors, \emph{Advances in Neural Information Processing Systems}, volume 38, Main Conference, pages 124505--124537. Curran Associates, Inc., 2025.
\newblock \doi{10.52202/085713-4150}.
\newblock URL \url{https://proceedings.neurips.cc/paper_files/paper/2025/file/b4689a8e6ac04c0919f7162365f1248d-Paper-Conference.pdf}.

\bibitem[Ma et~al.(2022)Ma, Xu, Sun, Yan, Zhang, and Ji]{x_clip}
Yiwei Ma, Guohai Xu, Xiaoshuai Sun, Ming Yan, Ji~Zhang, and Rongrong Ji.
\newblock X-clip: End-to-end multi-grained contrastive learning for video-text retrieval.
\newblock In \emph{Proceedings of the 30th ACM International Conference on Multimedia}, MM '22, page 638–647, New York, NY, USA, 2022. Association for Computing Machinery.
\newblock ISBN 9781450392037.
\newblock \doi{10.1145/3503161.3547910}.
\newblock URL \url{https://doi.org/10.1145/3503161.3547910}.

\bibitem[Morgado et~al.(2021)Morgado, Misra, and Vasconcelos]{AVID}
Pedro Morgado, Ishan Misra, and Nuno Vasconcelos.
\newblock Robust audio-visual instance discrimination.
\newblock In \emph{Proceedings of the IEEE/CVF Conference on Computer Vision and Pattern Recognition (CVPR)}, pages 12934--12945, June 2021.

\bibitem[Mu et~al.(2022)Mu, Kirillov, Wagner, and Xie]{SLIP}
Norman Mu, Alexander Kirillov, David Wagner, and Saining Xie.
\newblock Slip: Self-supervision meets language-image pre-training.
\newblock In \emph{Computer Vision – ECCV 2022: 17th European Conference, Tel Aviv, Israel, October 23–27, 2022, Proceedings, Part XXVI}, page 529–544, Berlin, Heidelberg, 2022. Springer-Verlag.
\newblock ISBN 978-3-031-19808-3.
\newblock \doi{10.1007/978-3-031-19809-0_30}.
\newblock URL \url{https://doi.org/10.1007/978-3-031-19809-0_30}.

\bibitem[Oord et~al.(2018)Oord, Li, and Vinyals]{InfoNCE}
Aaron van~den Oord, Yazhe Li, and Oriol Vinyals.
\newblock Representation learning with contrastive predictive coding.
\newblock \emph{arXiv preprint arXiv:1807.03748}, 2018.

\bibitem[Parida et~al.(2026)Parida, Emara, Doughty, and Damen]{Parida_2026_CVPR}
Kranti Parida, Omar Emara, Hazel Doughty, and Dima Damen.
\newblock Segmenting collision sound sources in egocentric videos.
\newblock In \emph{Proceedings of the IEEE/CVF Conference on Computer Vision and Pattern Recognition (CVPR) Workshops}, 2026.

\bibitem[Paul et~al.(2026)Paul, Patterson, and Bouguila]{Paul_2026_CVPR}
Sneha Paul, Zachary Patterson, and Nizar Bouguila.
\newblock Point cloud as a foreign language for multi-modal large language model.
\newblock In \emph{Proceedings of the IEEE/CVF Conference on Computer Vision and Pattern Recognition (CVPR)}, 2026.

\bibitem[Radford et~al.(2021)Radford, Kim, Hallacy, Ramesh, Goh, Agarwal, Sastry, Askell, Mishkin, Clark, Krueger, and Sutskever]{CLIP}
Alec Radford, Jong~Wook Kim, Chris Hallacy, Aditya Ramesh, Gabriel Goh, Sandhini Agarwal, Girish Sastry, Amanda Askell, Pamela Mishkin, Jack Clark, Gretchen Krueger, and Ilya Sutskever.
\newblock Learning transferable visual models from natural language supervision.
\newblock In \emph{International Conference on Learning Representations (ICLR)}, 2021.
\newblock \doi{10.48550/ARXIV.2103.00020}.
\newblock URL \url{https://arxiv.org/abs/2103.00020}.

\bibitem[Rao et~al.(2025)Rao, Lin, Chen, Xie, Zheng, and Yang]{11095210}
Jiahua Rao, Hanjing Lin, Leyu Chen, Jiancong Xie, Shuangjia Zheng, and Yuedong Yang.
\newblock Multi-modal contrastive learning with negative sampling calibration for phenotypic drug discovery.
\newblock In \emph{2025 IEEE/CVF Conference on Computer Vision and Pattern Recognition (CVPR)}, pages 30752--30762, 2025.
\newblock \doi{10.1109/CVPR52734.2025.02864}.

\bibitem[Sharma et~al.(2018)Sharma, Ding, Goodman, and Soricut]{sharma-etal-2018-conceptual}
Piyush Sharma, Nan Ding, Sebastian Goodman, and Radu Soricut.
\newblock Conceptual captions: A cleaned, hypernymed, image alt-text dataset for automatic image captioning.
\newblock In Iryna Gurevych and Yusuke Miyao, editors, \emph{Proceedings of the 56th Annual Meeting of the Association for Computational Linguistics (Volume 1: Long Papers)}, pages 2556--2565, Melbourne, Australia, July 2018. Association for Computational Linguistics.
\newblock \doi{10.18653/v1/P18-1238}.
\newblock URL \url{https://aclanthology.org/P18-1238/}.

\bibitem[Shvetsova et~al.(2022)Shvetsova, Chen, Rouditchenko, Thomas, Kingsbury, Feris, Harwath, Glass, and Kuehne]{everything_at_once}
Nina Shvetsova, Brian Chen, Andrew Rouditchenko, Samuel Thomas, Brian Kingsbury, Rogerio Feris, David Harwath, James Glass, and Hilde Kuehne.
\newblock Everything at once--multi-modal fusion transformer for video retrieval.
\newblock In \emph{2022 IEEE/CVF Conference on Computer Vision and Pattern Recognition (CVPR)}, pages 19988--19997. IEEE, 2022.

\bibitem[Srinivasa et~al.(2023)Srinivasa, Cho, Yang, Saidutta, Lee, Shen, and Jin]{CWCL}
Rakshith~Sharma Srinivasa, Jaejin Cho, Chouchang Yang, Yashas~Malur Saidutta, Ching-Hua Lee, Yilin Shen, and Hongxia Jin.
\newblock Cwcl: Cross-modal transfer with continuously weighted contrastive loss.
\newblock \emph{Advances in Neural Information Processing Systems}, 36:\penalty0 78496--78513, 2023.

\bibitem[Tian et~al.(2018)Tian, Shi, Li, Duan, and Xu]{ave}
Yapeng Tian, Jing Shi, Bochen Li, Zhiyao Duan, and Chenliang Xu.
\newblock Audio-visual event localization in unconstrained videos.
\newblock In \emph{ECCV}, 2018.

\bibitem[Tian et~al.(2019)Tian, Krishnan, and Isola]{CMC}
Yonglong Tian, Dilip Krishnan, and Phillip Isola.
\newblock Contrastive multiview coding.
\newblock In \emph{European Conference on Computer Vision (ECCV)}, 2019.
\newblock \doi{10.48550/ARXIV.1906.05849}.
\newblock URL \url{https://arxiv.org/abs/1906.05849}.

\bibitem[Wang et~al.(2024)Wang, Li, Li, Yu, He, Chen, Pei, Zheng, Wang, Shi, Jiang, Li, Xu, Zhang, Huang, Qiao, Wang, and Wang]{InternVideo2}
Yi~Wang, Kunchang Li, Xinhao Li, Jiashuo Yu, Yinan He, Guo Chen, Baoqi Pei, Rongkun Zheng, Zun Wang, Yansong Shi, Tianxiang Jiang, Songze Li, Jilan Xu, Hongjie Zhang, Yifei Huang, Yu~Qiao, Yali Wang, and Limin Wang.
\newblock Internvideo2: Scaling foundation models for multimodal video understanding.
\newblock page 396–416, Berlin, Heidelberg, 2024. Springer-Verlag.
\newblock ISBN 978-3-031-73012-2.
\newblock \doi{10.1007/978-3-031-73013-9_23}.
\newblock URL \url{https://doi.org/10.1007/978-3-031-73013-9_23}.

\bibitem[Wen et~al.(2026)Wen, Dai, Liu, Zheng, Dai, Wang, Kang, Wang, Xu, and Duan]{infMasking}
Liangjian Wen, Qun Dai, Jianzhuang Liu, Jiangtao Zheng, Yong Dai, Dongkai Wang, Zhao Kang, Jun Wang, Zenglin Xu, and Jiang Duan.
\newblock Infmasking: Unleashing synergistic information by contrastive multimodal interactions.
\newblock \emph{Advances in Neural Information Processing Systems}, 38:\penalty0 15529--15555, 2026.

\bibitem[Wu et~al.(2025)Wu, Qi, Yiling~Wu, Wang, and Wang]{yue25finecvr}
Yue Wu, Zhaobo Qi, Junshu~Sun Yiling~Wu, Yaowei Wang, and Shuhui Wang.
\newblock Learning fine-grained representations through textual token disentanglement in composed video retrieval.
\newblock \emph{ICLR}, 2025.

\bibitem[Wu et~al.(2023)Wu, Chen, Zhang, Hui, Berg-Kirkpatrick, and Dubnov]{CLAP}
Yusong Wu, Ke~Chen, Tianyu Zhang, Yuchen Hui, Taylor Berg-Kirkpatrick, and Shlomo Dubnov.
\newblock Large-scale contrastive language-audio pretraining with feature fusion and keyword-to-caption augmentation.
\newblock In \emph{IEEE International Conference on Acoustics, Speech and Signal Processing, ICASSP}, 2023.

\bibitem[Xu et~al.(2016)Xu, Mei, Yao, and Rui]{msrvtt}
Jun Xu, Tao Mei, Ting Yao, and Yong Rui.
\newblock Msr-vtt: A large video description dataset for bridging video and language.
\newblock In \emph{2016 IEEE Conference on Computer Vision and Pattern Recognition (CVPR)}, pages 5288--5296, 2016.
\newblock \doi{10.1109/CVPR.2016.571}.

\bibitem[Yuan et~al.(2021)Yuan, Lin, Kuen, Zhang, Wang, Maire, Kale, and Faieta]{cross_self}
Xin Yuan, Zhe Lin, Jason Kuen, Jianming Zhang, Yilin Wang, Michael Maire, Ajinkya Kale, and Baldo Faieta.
\newblock Multimodal contrastive training for visual representation learning.
\newblock In \emph{Proceedings of the IEEE/CVF Conference on Computer Vision and Pattern Recognition (CVPR)}, pages 6995--7004, June 2021.

\bibitem[Zhou et~al.(2024)Zhou, Li, Shang, Jiang, Liu, and Chen]{https://doi.org/10.48550/arxiv.2212.07699}
Jiawei Zhou, Xiaoguang Li, Lifeng Shang, Xin Jiang, Qun Liu, and Lei Chen.
\newblock Retrieval-based disentangled representation learning with natural language supervision.
\newblock In \emph{International Conference on Learning Representations (ICLR)}, 2024.
\newblock \doi{10.48550/ARXIV.2212.07699}.
\newblock URL \url{https://arxiv.org/abs/2212.07699}.

\end{thebibliography}
\bibliographystyle{plainnat}

\end{document}